\documentclass{article}

\usepackage{dblfloatfix}
\usepackage{float}
\usepackage{microtype}
\usepackage{graphicx}
\usepackage{subfigure}
\usepackage{booktabs} 
\usepackage{wrapfig}
\usepackage{natbib}
\usepackage{url}
\usepackage{array}
\usepackage{subcaption}
\usepackage{multirow}
\usepackage[utf8]{inputenc}
\usepackage{enumitem}
\usepackage{placeins}
\usepackage[normalem]{ulem}
\useunder{\uline}{\ul}{}

\usepackage{hyperref}

\newcommand{\revise}[1]{{\color{black} #1}}

\usepackage[accepted]{icml2025}

\usepackage{amsmath}
\usepackage{amssymb}
\usepackage{mathtools}
\usepackage{amsthm}

\usepackage[capitalize,noabbrev]{cleveref}

\theoremstyle{plain}

\theoremstyle{definition}

\theoremstyle{remark}

\usepackage[textsize=tiny]{todonotes}

\icmltitlerunning{UniCO: Towards a Unified Model for Combinatorial Optimization Problems}

\begin{document}

\twocolumn[
\icmltitle{UniCO: Towards a Unified Model for Combinatorial Optimization Problems}



\icmlsetsymbol{equal}{*}

\begin{icmlauthorlist}
\icmlauthor{Zefang Zong}{equal,yyy}
\icmlauthor{Xiaochen Wei}{equal,yyy}
\icmlauthor{Guozhen Zhang}{yyy}
\icmlauthor{Chen Gao}{yyy}
\icmlauthor{Huandong Wang}{yyy}
\icmlauthor{Yong Li}{yyy}

\end{icmlauthorlist}

\icmlaffiliation{yyy}{Department of Electronic Engineering, Tsinghua University, Beijing, China}

\icmlcorrespondingauthor{Zefang Zong}{zongzf19@mails.tsinghua.edu.cn}
\icmlcorrespondingauthor{Xiaochen Wei}{wwxc971231@163.com}

\icmlkeywords{Machine Learning, ICML}

\vskip 0.3in
]



\printAffiliationsAndNotice{\icmlEqualContribution} 

\begin{abstract}

Combinatorial Optimization (CO) encompasses a wide range of problems that arise in many real-world scenarios. While significant progress has been made in developing learning-based methods for specialized CO problems, a unified model with a single architecture and parameter set for diverse CO problems remains elusive. Such a model would offer substantial advantages in terms of efficiency and convenience. In this paper, we introduce UniCO, a unified model for solving various CO problems. Inspired by the success of next-token prediction, we frame each problem-solving process as a Markov Decision Process (MDP), tokenize the corresponding sequential trajectory data, and train the model using a transformer backbone. To reduce token length in the trajectory data, we propose a CO-prefix design that aggregates static problem features. To address the heterogeneity of state and action tokens within the MDP, we employ a two-stage self-supervised learning approach. In this approach, a dynamic prediction model is first trained and then serves as a pre-trained model for subsequent policy generation. Experiments across 10 CO problems showcase the versatility of UniCO, emphasizing its ability to generalize to new, unseen problems with minimal fine-tuning, achieving even few-shot or zero-shot performance. Our framework offers a valuable complement to existing neural CO methods that focus on optimizing performance for individual problems.

\end{abstract}

\section{Introduction}

Combinatorial optimization (CO) problems are pivotal in a wide range of real-world applications, including logistics and industrial management~\citep{singh2022combinatorial}. To address these generally NP-hard problems, traditional integer programming and heuristic methods have been extensively studied to obtain either exact or near-optimal solutions over the past decades. With the rapid growth of deep learning, solving CO problems using learning-based methods has garnered increasing attention, giving rise to the field of Neural Combinatorial Optimization (NCO)~\citep{kim2022sym,drakulic2024bq}.  \revise{Among all NCO schemes, the auto-regressive construction methods are favored in recent literature~\citep{bello2016neural, kool2018attention, kwon2020pomo, kim2022sym}. These methods construct solutions incrementally, and the entire problem-solving process can naturally be framed as a Markov Decision Process (MDP). These end-to-end methods offer significant computational efficiency and flexibility in generating feasible solutions, as they can easily avoid constraint-violating actions within the MDP framework~\citep{kim2022sym}.}


\begin{figure}
    \centering
        \includegraphics[width=0.35\textwidth]{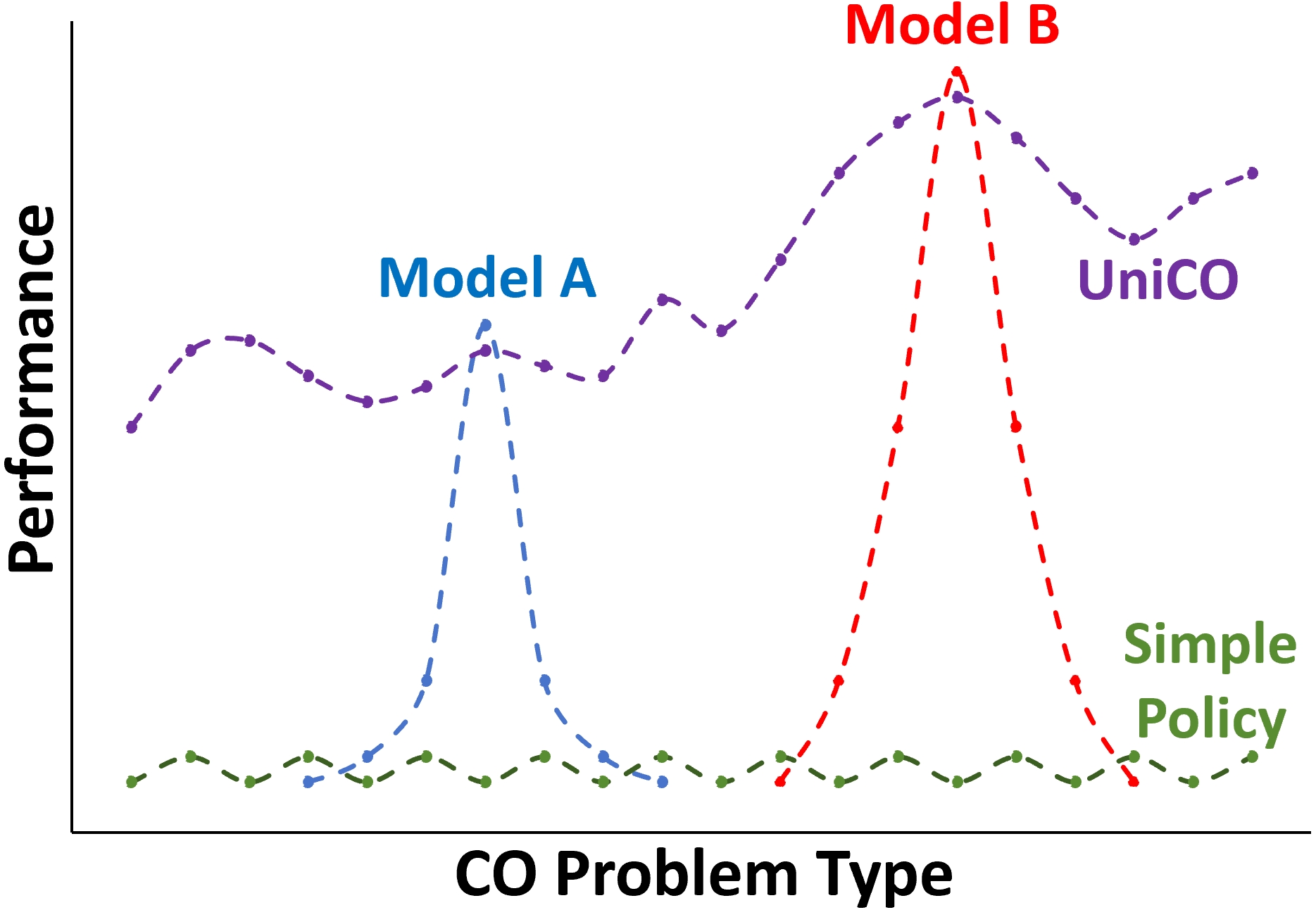}
        \vspace{-2mm}
        \caption{The No Free Lunch Theorem of optimization. 
        }
    \vspace{-4mm}
    \label{fig:lunch}
\end{figure}

\begin{figure*}
    \centering
    \includegraphics[width=0.85\linewidth]{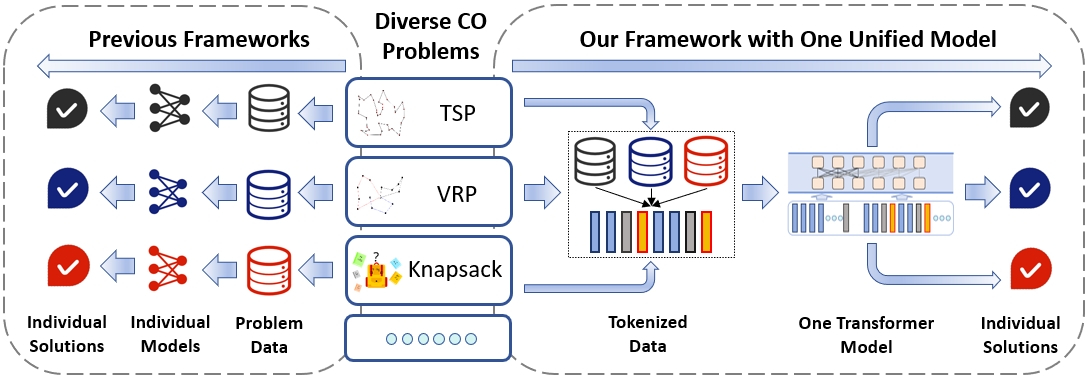}
    \vspace{-2mm}
    \caption{The difference between previous frameworks and ours to solve diverse CO problems. While previous frameworks require individual models with specific designs to adapt to different problems, our framework only utilizes one unified model.}
    \label{fig:overview}
    \vspace{-3mm}
\end{figure*}

However, a significant limitation remains: models from existing literature are typically tailored to specific problem types, lacking the ability to handle a wide range of problems simultaneously. There are clear advantages to using a unified model across diverse problems. First, it reduces the need for hand-crafted network structure designs for each individual problem. Second, it facilitates adaptation to unseen problem types more quickly and efficiently than training specific models from scratch. Although some literature claims to propose generic frameworks~\cite{kool2018attention,bello2016neural}, these methods generally apply the same general architecture to different problems, but with specific model structures and varying learning parameters, which results in a loss of true generality.
 The development of these NCO methods aligns with the famous No Free Lunch Theorem (NFLT)~\citep{wolpert1997no}. Most literature avoids the challenge to achieve generality across different problems, and focuses on improving performances on individual ones, illustrated as Model A and Model B in Figure~\ref{fig:lunch}. In contrast, we tackle this challenge across diverse CO problems, posing a new research question: Can we develop a unified model with a single neural architecture and parameter set that can simultaneously solve diverse CO problems, while maintaining strong few-shot capabilities?

Recently, the concept of next-token-prediction has marked a new era in general artificial intelligence, excelling in processing data across multiple scenarios, domains, and even modalities. The most successful examples are the large language models (LLMs) and multimodal large languange models (MLLMs) ~\citep{achiam2023gpt,dubey2024llama}, which can generalize across various natural language process (NLP) and computer vision (CV) scenarios and excel in few-shot learning tasks. Furthermore, the concept has also been applied to decision-making tasks directly~\citep{chen2021decision}. For instance, a generalist agent was developed to handle different control environments simultaneously, such as Atari games and robot benchmarks~\cite{reed2022generalist}. Motivated by these breakthroughs, we explore whether a unified model can be designed to tackle diverse CO problems under the same next-token-prediction framework. 




In general, we collect solutions for raw problem instances generated by state-of-the-art solvers from a variety of problem sources. Adopting the widely used auto-regressive MDP formulation from existing literature, we generate optimization trajectories where actions are iteratively selected based on partial solutions. These trajectories are serialized into flat token sequences and trained using a single transformer backbone, as illustrated in Figure~\ref{fig:overview}. However, directly applying existing training schemes to CO problems often proves inefficient. Since most CO problems are NP-hard, the observation space can be large, resulting in long token sequences and reduced training efficiency. Furthermore, a full trajectory contains different types of elements, including states and actions. Predicting all elements in a unified manner, without addressing their distinct roles and the heterogeneity between them, further complicates the training process.


To tackle these challenges, we propose the framework of UniCO incorporating two approaches to improve generic training performances considering the common characteristics of CO problems. First, we propose a non-causal, decoder-only architecture that incorporates a CO-prefix to reduce the overall token length. Unlike other environments where observations in an MDP can be fully dynamic, most information in a CO problem comes from its static description data. For example, in Traveling Salesman Problem (TSP), the distances between node pairs remain unchanged regardless of the visiting order. Therefore, we utilize a CO-prefix to aggregate the problems' static features, while the subsequent main trajectory handles dynamic observations. This reduces token length and improves training efficiency. Second, we decompose the entire token generation process into two self-supervised learning stages to reduce training difficulty. In the first stage, the model focuses solely on learning to predict forward dynamics, which then serves as the pre-trained model for the subsequent policy generation. These two stages are designed to handle the heterogeneous elements within the trajectory, thereby reducing the overall training difficulty.


While one recent work claims to have developed a generalist agent for combinatorial optimization problems~\cite{drakulic2024goal}, its applicability is limited to specific problem types. The approach relies on a specialized network backbone tailored to handle graph-based problems, but it does not extend to a broader range beyond such limit. In contrast, our UniCO can be applied to any CO problems as long as a feasible solution can be formulated as an MDP, which is a common property in CO.
 
To summarize, our key contributions are:
\begin{itemize}[leftmargin=*, itemsep=0pt, topsep=0pt, parsep=0pt, partopsep=0pt]
    \item We conduct an in-depth exploration of solving multiple combinatorial optimization (CO) problems using a single unified model, without the constraints of specific problem types.  We believe this approach provides a valuable complement to existing NCO methods that focus on achieving optimal performance for individual CO problems.
    \item  To overcome the challenges of directly applying traditional next-token prediction methods to CO problems, we introduce UniCO, a novel framework that incorporates a CO-prefix design and a two-stage self-supervised learning scheme. This approach effectively reduces token length and mitigates training complexities.
    \item We establish a comprehensive testbed featuring 10 CO problems to evaluate the generic problem-solving ability of our unified CO model. Experiments show that the model exhibits strong generic problem-solving capabilities. Additionally, we demonstrate its few-shot and even zero-shot generalization abilities when tackling new problems, enabled by fast fine-tuning.
\end{itemize}

\section{Related Works}
\subsection{Learning-based Methods for CO}

Solving CO problems via learning methods have drawn great attention recently, where many auto-regressive methods are highlighted. The pioneering work in this area was the Pointer Network, which was first tested on TSP\citep{vinyals2015pointer}. Subsequent research extended this idea by incorporating reinforcement learning (RL), demonstrating its effectiveness across a broader range of CO problems~\citep{bello2016neural}. Routing problems, a significant subclass of CO problems, have been extensively studied within this auto-regressive framework using RL~\citep{kool2018attention, kwon2020pomo}. To better account for both node and edge level features, a matrix-encoding framework was developed~\citep{kwon2021matrix}. The potential of applying auto-regressive NCO methods to more general CO problems was also discussed~\citep{drakulic2024bq}. These methods offer significant advantages due to their fast inference speed, as their computational complexity during testing remains low. Additionally, they are much more flexible in generating feasible actions that respect various problem constraints.

A recent trend in NCO research is exploring the generalization capabilities of algorithms. Existing methods primarily focus on generalizing across different data distributions~\citep{zhou2023towards, bi2022learning} and problem scales~\citep{zong2022rbg, li2021learning}. In terms of generalization to multiple problems, one study attempts to solve various VRPs by decomposing them into several elementary tasks~\citep{liu2024multi}. However, this decomposition relies heavily on human-designed rules, which limits its generalization potential. In contrast, we seek to develop a unified framework without specific problem type limitations.

\subsection{Next-Token-Prediction in Decision-Making}
In addition to the significant success of next-token prediction in both LLMs and MLLMs, researchers have also explored how to directly incorporate it into decision-making problems.
\cite{chen2021decision} first explored the use of the Transformer~\citep{vaswani2017attention} as an effective backbone for handling various control environments in an offline RL setting, including Atari, OpenAI Gym, and others. They trained a single policy model to generate actions at each step. \cite{janner2021offline} further proposed the Trajectory Transformer, which predicts all elements within a trajectory. In addition to offline RL, similar architectures have been integrated with imitation learning~\citep{reed2022generalist, shafiullah2022behavior, brohan2022rt, zhou2022policy}. A notable application of this approach is the Generalist Agent, GATO~\citep{reed2022generalist}, which successfully extended its capabilities across multiple control environments using a unified model. \cite{wen2022realization} further adapted the GATO structure, referred to as DB1, and extended it to solve TSP problems. Building on these successes, it is natural to consider Transformers as the backbone for a unified model capable of solving diverse CO problems. 

However, we note that \cite{wen2022realization} employed a pretrained GCN model~\citep{kipf2016semi} specifically trained for TSP to generate TSP state embeddings, rather than using the original TSP data directly. We believe this approach contradicts the core concept of a unified model, which should rely solely on a single architecture and parameter set. Nevertheless, we adopt the unified model structure proposed by GATO and re-implemented by DB1 as a key baseline for comparison, where only the original trajectory data is processed.

\section{Methodology}
In this section, we first introduce how diverse CO problems could be formulated and processed into a unified scheme for further training. Then we introduce the proposed UniCO framework, incorporating the non-causal transformer with CO-prefix design and the two-stage self-supervised learning.
\subsection{Data Preparation}
\subsubsection{Auto-regressive MDP Formulation for CO Problem}
We first formulate the sequential construction process of a CO problem solution as an MDP. Following the approach of existing auto-regressive NCO methods~\citep{zhang2023let}, a complete solution is incrementally constructed through multiple decision steps.

Let $\mathcal{S}$ denote the entire state space, with states $s_t \in \mathcal{S}$, and let $\mathcal{A} \subseteq \mathcal{S} \times \mathcal{S}$ be the action space, where actions are denoted by $a_t \in \mathcal{A}$. All states are assumed to be reachable from the initial state $s_1$. Since a CO problem is fully observed and deterministic, the transition from state $s_t$ to $s_{t+1}$ is fully determined by action $a_t$. Each state $s_t$  is represented as a set of actions taken before. A policy in the MDP refers to a distribution $P(s'|s)$ over the states $s'$ that can be reached from from $s$ via a single action. A feasible CO problem solution, represented as a complete trajectory $\tau$, can be further induced by the policy over $T$ steps via $\prod_{t=1}^{T}P(s_{t+1}|s_t)$.

It is important to note that \textit{tail recursion} is a common property in CO problems: after applying a series of construction steps, the remaining tail subproblem becomes a smaller instance of the original CO problem, as discussed in~\cite{drakulic2024bq}. It also includes in particular the Optimality Principle of Dynamic Programming~\cite{bellman1954theory}. Any tail-recursive problem can be formulated as the MDP described above, which enables us to further generate the trajectory dataset introduced later. 

\textbf{Proposision 1.} \textit{Any tail-recursive CO problem can be formulated as an MDP: states representing partial solutions, actions corresponding to decisions, transitions reflecting problem reduction, and rewards corresponding to objective values.}

\subsubsection{Trajectory Datasets}
To prepare the trajectory datasets for training, we first obtain the final optimized solutions from state-of-the-art solvers for various problems. We then trace their complete optimization MDP episodes, $\tau = (\tau_1, \tau_2, ..., \tau_T)$, where each episode consists of states and actions, with ${\tau_t} = (s_t, a_t)$ representing the state-action pairs at each step.

To jointly handle diverse features from different problems and distributions, we flatten all elements within the MDP episode into one dimension and tokenize them through a tokenization process. Discrete values, such as the node indices of actions, are directly assigned with integer token IDs from $[Min_d, Max_d)$. Continuous values, such as demands and positions, are first encoded via mu-law, discretized to $N_{bin}$ uniform bins, and then tokenized into the range $[Min_c, Max_c)$. The final trajectory token sequence $\overline{\tau}$ at each step is formulated with state tokens, followed by an action spliter token \texttt{<|>}, and then action tokens:

\begin{equation}
\vspace{-2mm}
    \overline{\tau} = (\overline{\tau_1}, \overline{\tau_2}, ..., \overline{\tau_T}), \quad \text{where} \, \overline{\tau_t}=(\overline{s_t}, \texttt{<|>}, \overline{a_t}).
\end{equation}

Note that the length of a fully tokenized sequence can sometimes be excessively long. To address this, we set the target total token length $L$ in advance, and use selected contiguous segments from complete solution MDPs. Additionally, we only preserve dynamic observations in the intermediate progress within $s_t$, while the static information of the raw problem instances is aggregated within a CO-prefix design, as introduced in the following section. For each problem instance and its complete solution MDP, we collect multiple trajectories as data augmentation. Details of tokenization and trajectory collection can be found in Appendix~\ref{appendix:tokenization}.

\begin{figure}[h]
\vspace{-2mm}
\centering
\subfigure[Causal decoder-only architecture without CO prefix.]{
\includegraphics[width=0.21\textwidth]{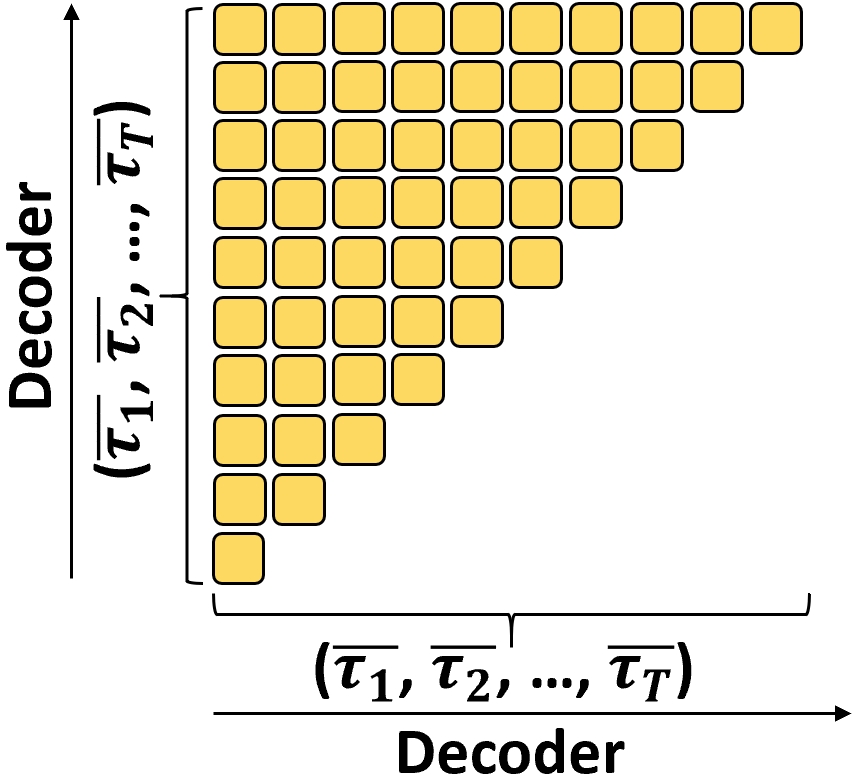}
\label{causal}
}
\hspace{0.01\textwidth}
\subfigure[Non-causal decoder-only architecture with CO-prefix.]{
\includegraphics[width=0.21\textwidth]{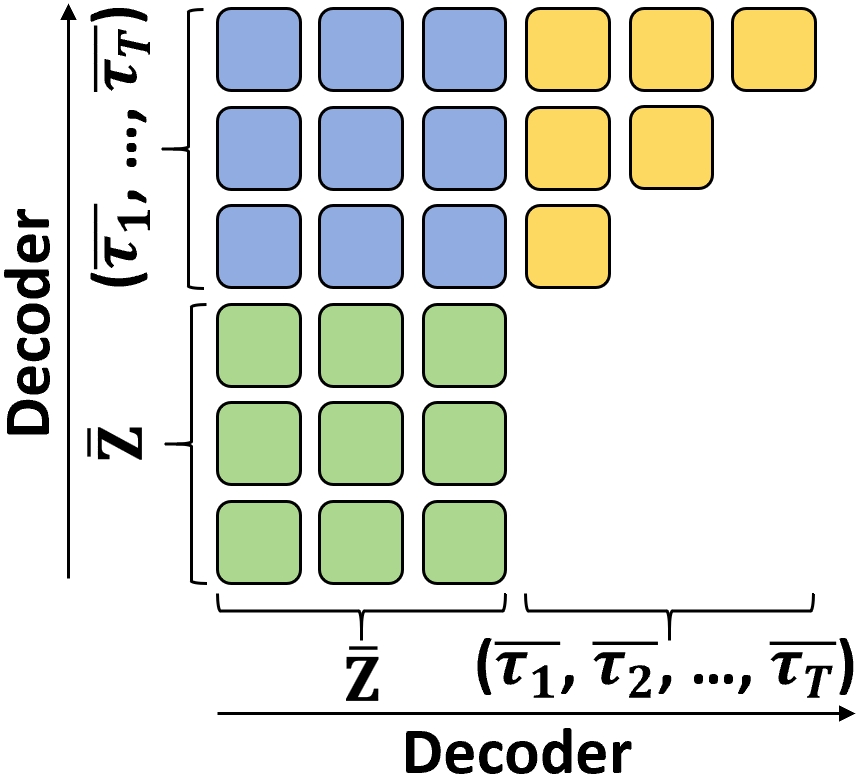}
\label{non-causal}
}
\vspace{-2mm}
\caption{Two architecture designs for a unified model. a) Causal decoder-only architecture without CO prefix, where each token is only conditioned on the past tokens and only trajectory data is processed, adopted in ~\cite{reed2022generalist}. The entire token length is large. b) Non-causal decoder-only architecture with CO-prefix, where tokens in the CO-prefix shares richer representations conditioned on both prior and past tokens. The trajectory no longer process duplicated static information.}
\label{network}
\vspace{-2mm}
\end{figure}

\begin{figure*}[h]
\centering
\subfigure[The dynamics forward stage.]{
\includegraphics[width=0.4\textwidth]{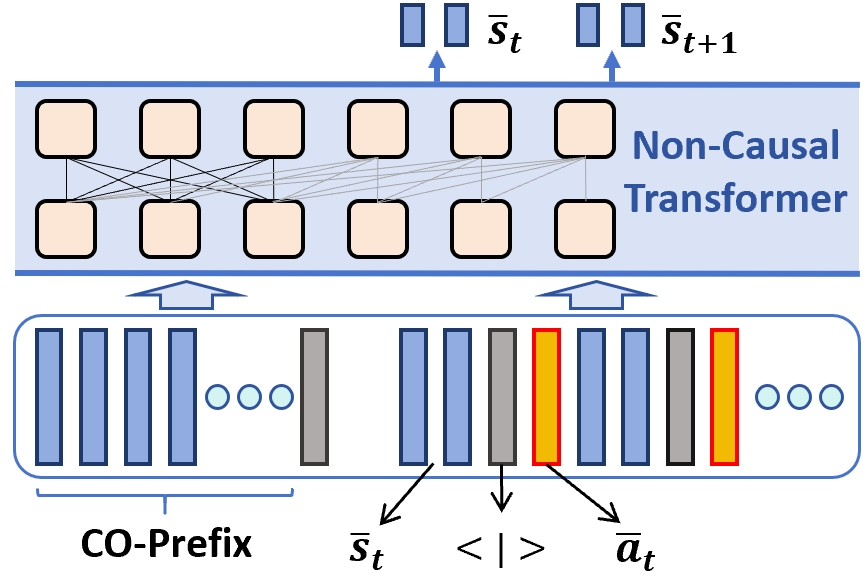}
\label{stage1}
}
\hspace{0.05\textwidth}
\subfigure[The policy generation stage.]{
\includegraphics[width=0.4\textwidth]{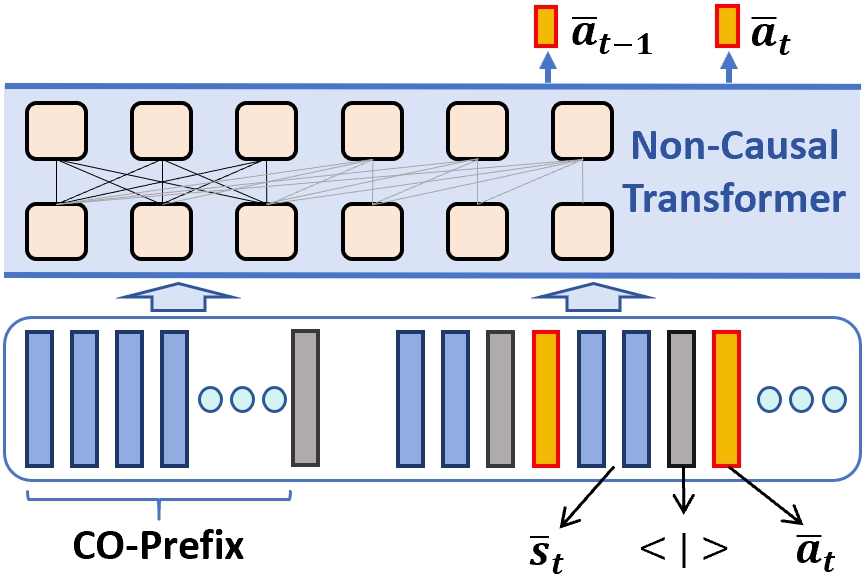}
\label{stage2}
}
\vspace{-4mm}
\caption{Two-stage self-supervised learning to train the unified CO model.}
\label{fig:framework}
\vspace{-4mm}
\end{figure*}

\subsection{Non-causal Transformer with CO-prefix}

Due to the NP-hard nature of most CO problems, the observation space and dimensionality can be large, resulting in long token sequences and reduced training efficiency.

To tackle this challenge, we decompose the original state representation into static and dynamic components, as most of the information in a CO problem comes from its static description data. For instance, in a TSP instance, the positions of the cities are static and remain unchanged throughout the optimization MDP, while the dynamic information only includes the current position. We further introduce a CO-prefix design to capture the static information, which is prepended to the beginning of the token trajectory. The subsequent sequence then focuses solely on dynamic observations. This approach avoids duplicating the representation of observations by tokenizing only the current dynamic state at each step, rather than the entire information. This design significantly reduces token length and improves training efficiency. Let $P$ and $\overline{P}$ represent the raw and tokenized CO-prefix, respectively. The final token sequence fed into the model is $(\overline{P}; \texttt{<X>}, \overline{\tau})$, where \texttt{<X>} denotes a separator token between them.

Although the sequential nature of Markov Decision Processes (MDPs) with time-dependent ordering makes the causal transformer architecture a natural choice due to its simple and effective one-directional design, as suggested in previous sequential decision-making literature~\citep{chen2021decision, reed2022generalist}, shown in Figure~\ref{causal}, it has certain limitations. \revise{Specifically, the CO-prefix $P$ is time-invariant, as it only contains static representations. Therefore, each token within $\overline{P}$ should be fully visible and processed with each other in a bi-directional manner.}

To address this, we adopt a non-causal transformer architecture, where CO-prefix tokens are processed bi-directionally to ensure comprehensive context integration, while the remainder of the sequence is handled in a one-directional manner, as shown in Figure~\ref{non-causal}. The CO-prefix tokens share richer representations, conditioned on both preceding and subsequent tokens, which enhances overall performance.

\textbf{Action and CO-prefix Mask} \quad To ensure that each action selected by UniCO is feasible during inference, the output policy must be masked to filter out actions that violate problem constraints, using the action mask provided by the problem environment. It is important to note that during the generation of trajectory data, action masks are collected alongside the trajectory data at each step. The action mask is transformed into the CO-prefix mask, where each token corresponding to an infeasible action is masked in the attention module. For example, in the Traveling Salesman Problem (TSP), the CO-prefix mask includes the coordinates of already visited cities. In the Flexible Flow Shop Problem (FFSP), it corresponds to the job duration entries of completed tasks. This design allows the model to focus on more relevant tokens for feasible actions, without increasing the overall token length.

While the CO-prefix design is efficient, UniCO is not strictly dependent on it and can seamlessly handle problems without such prefix information, including fully dynamic problems. This capability is demonstrated in our evaluation results. Overall, UniCO is not constrained by specific problem types.

\subsection{Two-Stage Self-Supervised Learning}

Since a complete trajectory consists of different types of elements, such as observations and actions, predicting them without distinguishing their individual roles further increases the training difficulty. 

\revise{To address this challenge, we decompose the token generation process into two stages in a self-supervised learning framework: a dynamics forward stage and a policy generation stage, as shown in Figure~\ref{fig:framework}.}

\begin{itemize}[leftmargin=*, itemsep=0pt, topsep=0pt, parsep=0pt, partopsep=0pt]
    \item \textbf{Dynamics forward stage.} In the first stage, we pre-train the model to predict the next observation given the current action. The training loss for a training batch $\mathcal{B}$ is defined as follows:
    \begin{equation}
    \resizebox{0.46\textwidth}{!}{$
        \mathcal{L}(\theta, \mathcal{B}) = -\sum_{b=1}^{|B|}\sum_{t=1}^{T^b}\text{log}p_{\theta}(\overline{s_{t+1}^b}|(\overline{P^b}, \texttt{<X>},\overline{\tau_{1:t}^b})),
        $}
    \end{equation}
    where $T^b$ is the amount of trajectory units in the current token length. Since MDP transitions are deterministic in CO problems, ths dynamics model can be accurately trained with the same amount of data. 
    \item \textbf{Policy generation stage.} In the second stage, we fine-tune the model to generate actions based on the pretrained model in advance. The training loss for a training batch $\mathcal{B}$ is defined as follows:
    \begin{equation}
        \resizebox{0.46\textwidth}{!}{$
        \mathcal{L}(\theta, \mathcal{B}) = -\sum_{b=1}^{|B|}\sum_{t=1}^{T^b} \text{log} p_{\theta}(\overline{a_{t+1}^b} \mid (\overline{P^b}, \texttt{<X>}, \overline{\tau_{1:t}^b}, \overline{s_{t+1}^b}, \texttt{<|>}))$
        }
    \end{equation}

\end{itemize}

 This two-stage decomposition simplifies the learning process by decomposing the overall process into two sub-tasks, allowing the model to first understand intermediate dynamics and then generate qualified policy. This leads to faster and more effective convergence during training.



\section{Performance Evaluation}

\subsection{Problem and Expert Selection}

To evaluate the generic problem-solving ability of UniCO, we construct a set of 10 diverse problems for assessment.

        

\begin{table}[h]

    \caption{The summary of the evaluated CO problems, including individual expert solver to collect trajectories, the prefix token length and the step state token length. $N$ denotes the number of nodes, items, or jobs, depending on the problem, and $M$ denotes the number of machines in the FFSP.}
    \label{tab:expert}
    \begin{center}
    \resizebox{0.48\textwidth}{!}{
    \begin{tabular}{c|c|c|c}
    \hline
        \textbf{Problem} & \textbf{Expert Solver} &\textbf{Prefix-Token} & \textbf{State-Token}  \\ 
        \hline
        TSP  & LKH3&$2N$&$2$ \\
        VRP  & LKH3&$3N+2$&$3$  \\
        OP   & Gurobi& $3N+2$&$4$ \\
        PCTSP & ILS\footnotemark &$4N+2$&$3$\\
        SPCTSP & ILS &$4N+2$&$3$ \\
        Knapsack & DP & $2N$ &$1$ \\
        ATSP &LKH3&$N\times N$&$N$\\
        MIS &Kamis&$N \times N$ &$N$\\
        FFSP &MatNet&$N\times M$&$M+1$ \\
        3DBP &PCT & - &$6\times(N+1)$ \\
    \hline
    \end{tabular}
    }
    \end{center}

\end{table}

\footnotetext{https://github.com/jordanamecler/PCTSP}

We first select four common routing problems that have been extensively studied in recent literature~\citep{kool2018attention, kim2022sym}, including Traveling Salesman Problem (TSP), Vehicle Routing Problem (VRP), Orienteering Problem (OP) and Prize Collecting TSP (PCTSP). To demonstrate how our model handles uncertainty, we also include the Stochastic PCTSP (SPCTSP). We also consider Asymmetric TSP (ATSP), where the problem is defined on adjacency matrix without Cartesian coordinates~\citep{kwon2021matrix}. Beyond routing problems, we evaluate our model on the Knapsack problem following previous NCO literature~\cite{bello2016neural, grinsztajn2023winner}. We also include the Maximum Independent Set (MIS) problem~\citep{sun2023difusco} and the Flexible Flow Shop Problem (FFSP)~\citep{kwon2020pomo}. Finally, to demonstrate the capability of UniCO on fully dynamic and non-graph based problems, we also include online 3D Bin-packing (3DBP)~\cite{zhao2021learning}.

For each problem, trajectories are collected from individual expert solver, as shown in Table~\ref{tab:expert}.The problem scale is set to $N=20$, where $N$ represents the number of nodes, items, or jobs, depending on the problem. The instance generation scheme is aligned with previous literature for each problem. Details of data generation and token design can be found in Appendix~\ref{problems}.

\begin{table*}[!htbp] 
\caption{
Performance results. The best learning-based results are underlined, and the best results of unified models are in bold.
}
\centering
\resizebox{0.88\textwidth}{!}{

\begin{tabular}{c|cccc|cccc}
\hline\hline
                  & \multicolumn{4}{c|}{TSP}                                                         & \multicolumn{4}{c}{Knapsack}                                                     \\
Method            & Obj.$\downarrow$       & Gap$\downarrow$          & Score$\uparrow$           & Time$\downarrow$ & Obj.$\uparrow$ & Gap $\downarrow$         & Score $\uparrow$          & Time$\downarrow$\\
\hline
Random            & 10.47                & -                      & 0.00\%                  & (9s)  & 38.14                 & -                      & 0.00\%                  & (6s)  \\
Expert            & 3.84                 & 0.00\%                 & 100.00\%                & (2h)  & 63.89                 & 0.00\%                 & 100.00\%                & (10m) \\
POMO-single traj  & 3.84                 & 0.07\%                 & 99.98\%                 & (22s) & 63.14                 & 1.17\%                 & 97.09\%                 & (30s) \\
POMO              & {\ul 3.84}           & {\ul 0.01\%}           & {\ul 99.99\%}           & (23s) & {\ul 63.79}           & {\ul 0.16\%}           & {\ul 99.61\%}           & (31s) \\
GATO/DB1-greedy   & 3.99                 & 3.80\%                 & 97.68\%                 & (1h)  & 62.19                 & 2.66\%                 & 93.40\%                 & (35m) \\
GATO/DB1-sampling & 3.86                 & 0.49\%                 & 99.70\%                 & (15h) & 63.53                 & 0.56\%                 & 98.60\%                 & (8h)  \\
UniCO-DR           & 3.88                 & 1.04\%                 & 99.40\%                 & (1m)  & 61.78                 & 3.30\%                 & 91.81\%                 & (51s) \\
UniCO-greedy       & 3.87                 & 0.78\%                 & 99.55\%                 & (1m)  & 61.99                 & 2.97\%                 & 92.62\%                 & (50s) \\
UniCO-samping      & {\ul \textbf{3.84}}  & {\ul \textbf{0.01\%}}  & {\ul \textbf{99.99\%}}  & (23m) & \textbf{63.56}        & \textbf{0.26\%}        & \textbf{98.72\%}        & (13m) \\
\hline\hline
                  & \multicolumn{4}{c|}{CVRP}                                                        & \multicolumn{4}{c}{OP}                                                           \\
Method            & Obj.$\downarrow$       & Gap $\downarrow$         & Score $\uparrow$          & Time$\downarrow$ & Obj.$\uparrow$          & Gap $\downarrow$         & Score  $\uparrow$         & Time$\downarrow$\\
\hline
Random            & 13.25                & -                      & 0.00\%                  & (29s) & 1.93                  & -                      & 0.00\%                  & (8s)  \\
Expert            & 6.11                 & 0.00\%                 & 100.00\%                & (5h)  & 5.38                  & 0.00\%                 & 100.00\%                & (1h)  \\
AM-greedy         & 6.38                 & 4.40\%                 & 96.12\%                 & (7s)  & 5.19                  & 3.72\%                 & 93.86\%                 & (9s)  \\
AM-sampling       & 6.29                 & 2.96\%                 & 97.40\%                 & (14m) & 5.26                  & 2.55\%                 & 95.78\%                 & (7m)  \\
GATO/DB1-greedy   & 6.63                 & 8.51\%                 & 92.72\%                 & (2h)  & 4.91                  & 8.87\%                 & 85.46\%                 & (53m) \\
GATO/DB1-sampling & 6.27                 & 2.41\%                 & 97.82\%                 & (18h) & 5.30                  & 1.56\%                 & 97.42\%                 & (10h) \\
UniCO-DR           & 6.75                 & 10.47\%                & 91.04\%                 & (2m)  & 5.00                  & 7.06\%                 & 88.99\%                 & (1m)  \\
UniCO-greedy       & 6.66                 & 9.00\%                 & 92.30\%                 & (2m)  & 5.06                  & 5.95\%                 & 90.72\%                 & (1m)  \\
UniCO-samping      & {\ul \textbf{6.27}}  & {\ul \textbf{2.40\%}}  & {\ul \textbf{97.85\%}}  & (31m) & {\ul \textbf{5.32}}   & {\ul \textbf{1.21\%}}  & {\ul \textbf{98.01\%}}  & (18m) \\
\hline\hline
                  & \multicolumn{4}{c|}{PCTSP}                                                       & \multicolumn{4}{c}{SPCTSP}                                                       \\
Method            & Obj.$\downarrow$       & Gap $\downarrow$         & Score $\uparrow$          & Time$\downarrow$ & Obj.$\downarrow$ & Gap  $\downarrow$      & Score $\uparrow$          & Time$\downarrow$\\
\hline
Random            & 9.25                 & -                      & 0.00\%                  & (20s) & 9.24                  & -                      & 0.00\%                  & (20s) \\
Expert            & 3.16                 & 0.00\%                 & 100.00\%                & (2h)  & 3.31                  & 0.00\%                 & 100.00\%                & (2h)  \\
AM-greedy         & 3.18                 & 0.85\%                 & 99.57\%                 & (13s) & 3.23                  & -0.71\%                & 101.25\%                & (9s)  \\
AM-sampling       & 3.16                 & 0.13\%                 & 99.97\%                 & (12m) & 3.20                  & -1.85\%                & 101.94\%                & (10m) \\
GATO/DB1-greedy   & 3.27                 & 3.48\%                 & 98.19\%                 & (1h)  & 3.30                  & -0.30\%                & 100.17\%                & (1h)  \\
GATO/DB1-sampling & 3.20                 & 1.26\%                 & 99.36\%                 & (15h) & 3.28                  & -0.90\%                & 100.47\%                & (16h) \\
UniCO-DR           & 3.27                 & 3.48\%                 & 98.19\%                 & (2m)  & 3.28                  & -0.91\%                & 100.51\%                & (2m)  \\
UniCO-greedy       & 3.20                 & 1.27\%                 & 99.34\%                 & (2m)  & 3.26                  & -1.51\%                & 100.84\%                & (2m)  \\
UniCO-samping      & {\ul \textbf{3.15}}  & {\ul \textbf{-0.27\%}} & {\ul \textbf{100.21\%}} & (26m) & {\ul \textbf{3.16}}   & {\ul \textbf{-4.03\%}} & {\ul \textbf{102.89\%}} & (27m) \\
\hline\hline
                  & \multicolumn{4}{c|}{ATSP}                                                        & \multicolumn{4}{c}{FFSP}                                                         \\
Method            & Obj. $\downarrow$      & Gap $\downarrow$         & Score $\uparrow$          & Time$\downarrow$ & Obj.$\downarrow$        & Gap $\downarrow$         & Score $\uparrow$          & Time $\downarrow$\\
\hline
Random            & 10.49                & -                      & 0.00\%                  & (10s) & 45.00                 & -                      & 0.00\%                  & (12m) \\
Expert            & 3.85                 & 0.00\%                 & 100.00\%                & (2h)  & 27.31                 & 0.00\%                 & 100.00\%                & (5m)  \\
MatNet            & 3.87                 & 0.52\%                 & 99.70\%                 & (33s) & {\ul 27.31}           & {\ul 0.00\%}           & {\ul 100.00\%}          & (5m)  \\
MatNet-augment    & {\ul 3.85}           & {\ul 0.03\%}           & {\ul 99.98\%}           & (7m)  & -                     & -                      & -                       & -     \\
GATO/DB1-greedy   & 10.47                & 171.95\%               & 0.30\%                  & (32m) & 41.42                 & 51.67\%                & 20.24\%                 & (4h)  \\
GATO/DB1-sampling & 8.86                 & 131.09\%               & 22.78\%                 & (8h)  & 41.01                 & 50.16\%                & 22.56\%                 & (65h) \\
UniCO-DR           & 4.38                 & 13.76\%                & 91.87\%                 & (2m)  & 29.20                 & 6.92\%                 & 89.32\%                 & (29m) \\
UniCO-greedy       & 4.22                 & 9.61\%                 & 94.43\%                 & (2m)  & 29.11                 & 6.59\%                 & 89.82\%                 & (27m) \\
UniCO-samping      & \textbf{3.96}        & \textbf{3.04\%}        & \textbf{98.15\%}        & (44m) & \textbf{28.34}        & \textbf{3.77\%}        & \textbf{94.18\%}        & (7h)  \\
\hline\hline
                  & \multicolumn{4}{c|}{MIS}                                                         & \multicolumn{4}{c}{3DBP}                                                         \\
Method            & Obj. $\uparrow$        & Gap $\downarrow$         & Score $\downarrow$        & Time$\downarrow$ & Obj.$\downarrow$ & Gap$\downarrow$ & Score $\uparrow$          & Time$\downarrow$  \\
\hline
Random            & 9.11                 & -                      & 0.00\%                  & (7s)& 0.1739                & -                      & 0.00\%                  & (20s)\\
Expert            & 10.44                & 0.00\%                 & 100.00\%                & (7m)  & 0.8182                & 0.00\%                 & 100.00\%                & (11m)\\
LwD-greedy        & 10.42                & 0.19\%                 & 98.50\%                 & (8m)  & -                     & -                      & -                       & -     \\
GATO/DB1-greedy   & 9.70                 & 7.09\%                 & 44.36\%                 & (33m) & 0.7729                & 5.54\%                 & 92.83\%                 & (2h)\\
GATO/DB1-sampling & 9.82                 & 5.94\%                 & 53.38\%                 & (8h)  & 0.7831                & 4.29\%                 & 94.55\%                 & (21h)\\
UniCO-DR           & 10.35                & 0.86\%                 & 93.23\%                 & (3m)  & 0.8101                & 0.99\%                 & 98.74\%                 & (41m)\\
UniCO-greedy       & 10.35                & 0.86\%                 & 93.23\%                 & (3m)  & 0.8123                & 0.72\%                 & 99.08\%                 & (41m)\\
UniCO-samping      & {\ul \textbf{10.40}} & {\ul \textbf{0.38\%}}  & {\ul \textbf{97.00\%}}  & (1h)  & {\ul \textbf{0.8152}} & {\ul \textbf{0.37\%}}  & {\ul \textbf{99.53\%}}  & (10h)\end{tabular}
}
\label{tab:main-result}
\end{table*}

\subsection{Evaluation Protocols}
\textbf{Hyperparameters} \quad During training, each epoch consists of 400 batches, with 128 \revise{trajectories} in each batch. The trajectory data for each epoch is newly sampled from a mixed set of all 10 problems. The total token length of each trajectory is $L=1000$, either clipped or padded from the complete MDP episode data concatenated to the CO-prefix. The transformer architecture uses 10 layers with 768 embedding dimensions. For tokenization, the discrete range is set to $[0,200)$, the continuous range to $[0,4]$, and the bin number to 1800. We evaluate the model on the validation dataset every two epochs and apply early stopping if no improvement is observed over 6 consecutive epochs. During inference, we use the \textit{KV-Cache} technique to accelerate problem solving. Performances are evaluated on each problem individually, using a test dataset of 10,000 instances per problem. Further implementation details are provided in Appendix~\ref{appendix:implement} for reproducibility.

\textbf{Metrics} \quad  We report four metrics respectively. Following previous NCO literature~\citep{kool2018attention}, we present the original objectives, the gap from expert results, and the evaluation time on the entire test dataset. Additionally, in line with literature on generic decision-making~\citep{reed2022generalist}, we report performance scores as a percentage, where 100\% represents the expert performance for each task, and 0\% corresponds to a random policy. The score is calculated as $Score = |obj_e-obj_r|/|obj-obj_r|$, where $obj_e$ and $obj_r$ denote the objectives of the expert and a random policy respectively~\citep{wen2022realization}. 

\textbf{Baselines} \quad We evaluate UniCO with two variations: with and without the two-stage supervised learning. We refer to the model directly trained to generate actions as \textit{UniCO-DR}, as shown in Table~\ref{tab:main-result}.
For baseline comparisons, we first demonstrate the corresponding expert approach for each problem as a straightforward benchmark. We then compare our model with GATO~\citep{reed2022generalist}, which was re-implemented and reported by~\cite{wen2022realization} as DB1.  Note that we manually implemented the original GATO framework, as it is not open-sourced. Unlike our approach, GATO is trained using a causal transformer structure, where the trajectory data for each problem is prepended with a prompt sequence from the same problem. The prompt consists of multiple step transitions from other episodes, and other key hyperparameters remain the same as ours. Both GATO and UniCO are evaluated using two decoding strategies: greedy decoding and sampling, with $16$ solutions per evaluation. Finally, we compare UniCO with auto-regressive specialist NCO methods, which also use the MDP formulation for CO problems. We report performance on the TSP and Knapsack problems for POMO~\citep{kwon2020pomo}, CVRP, OP, PCTSP, and SPCTSP for AM~\citep{kool2018attention}, ATSP and FFSP for MatNet~\citep{kwon2021matrix}, MIS for LwD~\citep{ahn2020learning} and 3DBP for PCT~\cite{zhao2021learning}. Note that MatNet and PCT are used as both the expert solver and the learning baseline for FFSP and 3DBP respectively. We also report performance for a random policy, along with the evaluation time, which reflects the environment time cost in our implementation.


\subsection{Performances of Generic Problem Solving}

The main evaluation results across all 10 problems are illustrated in Table~\ref{tab:main-result}. The best learning-based results, whether specialist or unified, are underlined, while the best results among all unified models are shown in bold. 


\textbf{UniCO demonstrates strong generic problem-solving abilities, achieving performance comparable to specialist models. } With 16-sample decoding, our model achieves scores above 97.00\% on all problems except FFSP. Remarkably, when using sampling, our model even outperforms specialist learning baselines under the same setting on 6 out of the 10 problems.

\textbf{The CO-prefix design is significant.} We found that GATO/DB1 struggles to converge effectively on ATSP, FFSP, and MIS, primarily due to its lack of a prefix design. Without this design, GATO/DB1 computes full observation tokens at each step, which becomes highly inefficient when the observation space is large. For instance, in both ATSP and MIS, the static information is represented by the instance adjacency matrix, which has a complexity of $O(N^2)$. In each training episode, GATO/DB1 can only process one or two complete trajectory steps, with or without their prepended prompt sequences. The sparse loss signals from action tokens hinder the model's convergence. 

\textbf{The two-stage self-supervised learning scheme improves performances.} Compared to UniCO that is directly trained to generate actions, a model fine-tuned on a pre-trained forward dynamics model outperforms across all 10 problems when evaluated with greedy decoding. The separation of dynamics prediction and action generation significantly reduces the overall training difficulty, leading to improved solution quality.


\subsection{Performances on Few-shot Ability}

\begin{figure}[h]
    \centering
    \includegraphics[width=0.44\textwidth]{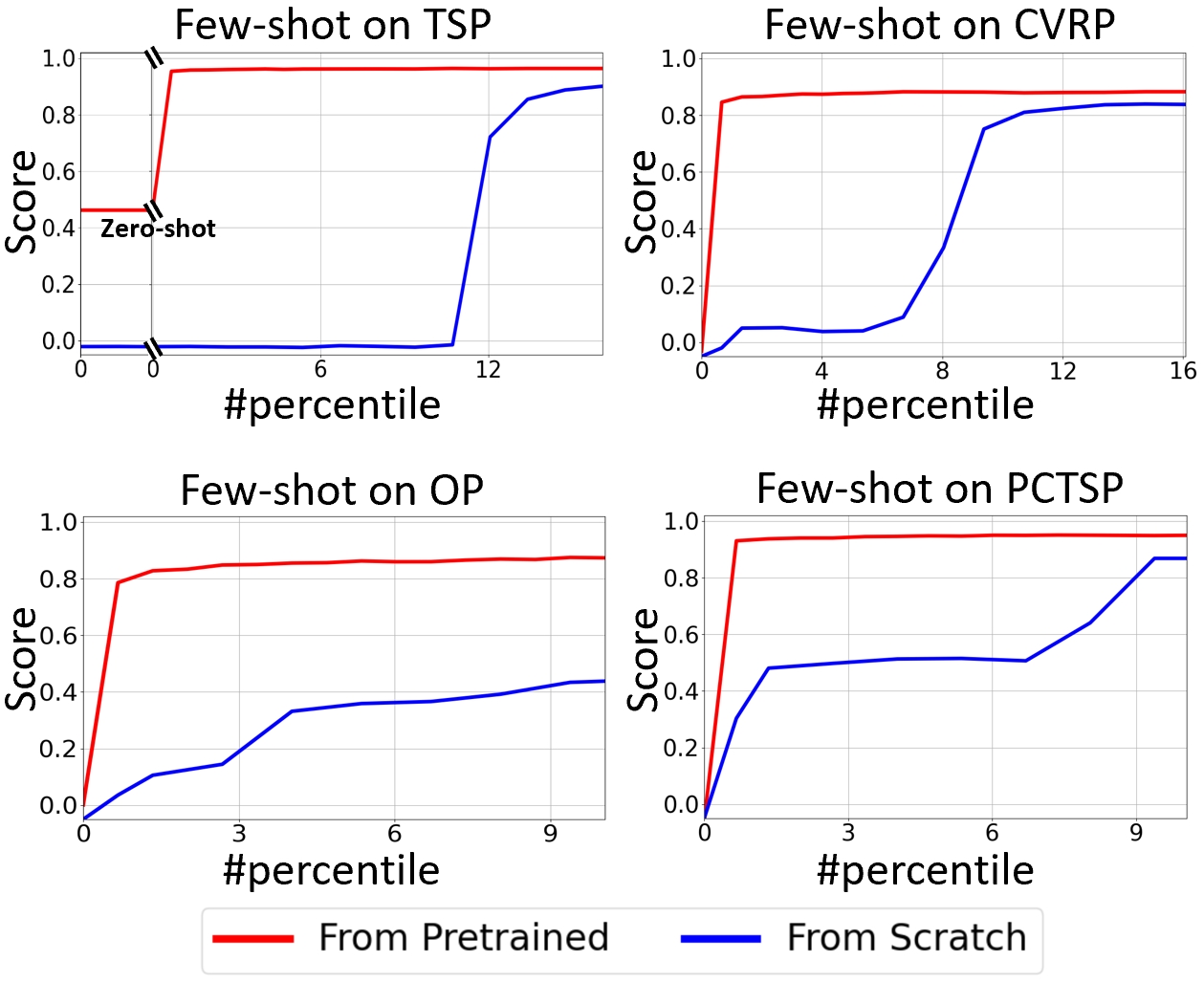}
    \caption{The few-shot results on four routing problems. The x-axis represents the percentage of data used for fine-tuning in relation to the data used in the main results.}
    \label{fig:few-shot}

\end{figure}

To evaluate the few-shot generalization ability of our model on unseen problems, we select four routing problems and train four distinct unified models. \revise{Each model is trained in a leave-one-out manner, excluding the selected problem, and then gradually fine-tuned using datasets from the unseen problem. In each epoch for fine-tuning, we use $0.67\%$ of the total data that was used for the problem in the main results.} We report the optimization scores and compare them with those of a model trained from scratch on the corresponding problem, \revise{as shown in Figure~\ref{fig:few-shot}}.

Overall, our model demonstrates strong \revise{few-shot generalization} across all four problem settings, even with limited data. In each case, the model achieves high solution quality after just one epoch with minimal data. These results show that our pre-trained unified model can be quickly adapted to an unseen problem, eliminating the need for time-consuming retraining of a separate model. This significantly enhances both convenience and efficiency, making it well-suited for real-world applications.

In addition to few-shot abilities, we observed even zero-shot generalization on TSP. The corresponding prefix and step token designs, which only include city coordinates, represent a subset of the more complex routing problems. Our pre-trained model, originally trained on these high-level problems, is able to directly generate solutions with approximately 48\% optimality without any additional fine-tuning data.

\subsection{Additional Properties of UniCO}
Besides the performances on generic problem solving and few-shot ability, we also demonstrate the versatile problem solving ability in larger problem scales in Appendix~\ref{scale-50}, and the strong generalization ability to different scales via fine-tuning. We also evaluate how different problem type combinations affect the overall performances in Appendix~\ref{prob-combination}, where we found that a mix of different problem types even further improve the overall performances. Finally, we conduct ablation studies on hyperparameters in Appendix~\ref{ablation}.

\section{Conclusion and Future Works}
\label{conclusion}
In this paper, we propose UniCO, a unified model to solve diverse CO problems simultaneously. We evaluated the performance of our proposed model on 10 different problems, demonstrating that our approach provides a valuable complement to existing NCO methods that focus on optimizing performance for individual CO problems. As for our future work, we plan to enhance our model to tackle problems with significantly longer token sequences, corresponding to industrial problems with much larger scales.

\bibliography{example_paper}

\begin{thebibliography}{41}
\providecommand{\natexlab}[1]{#1}
\providecommand{\url}[1]{\texttt{#1}}
\expandafter\ifx\csname urlstyle\endcsname\relax
  \providecommand{\doi}[1]{doi: #1}\else
  \providecommand{\doi}{doi: \begingroup \urlstyle{rm}\Url}\fi

\bibitem[Achiam et~al.(2023)Achiam, Adler, Agarwal, Ahmad, Akkaya, Aleman, Almeida, Altenschmidt, Altman, Anadkat, et~al.]{achiam2023gpt}
Achiam, J., Adler, S., Agarwal, S., Ahmad, L., Akkaya, I., Aleman, F.~L., Almeida, D., Altenschmidt, J., Altman, S., Anadkat, S., et~al.
\newblock Gpt-4 technical report.
\newblock \emph{arXiv preprint arXiv:2303.08774}, 2023.

\bibitem[Ahn et~al.(2020)Ahn, Seo, and Shin]{ahn2020learning}
Ahn, S., Seo, Y., and Shin, J.
\newblock Learning what to defer for maximum independent sets.
\newblock In \emph{International conference on machine learning}, pp.\  134--144. PMLR, 2020.

\bibitem[Balas(1989)]{balas1989prize}
Balas, E.
\newblock The prize collecting traveling salesman problem.
\newblock \emph{Networks}, 19\penalty0 (6):\penalty0 621--636, 1989.

\bibitem[Bellman(1954)]{bellman1954theory}
Bellman, R.
\newblock The theory of dynamic programming.
\newblock \emph{Bulletin of the American Mathematical Society}, 60\penalty0 (6):\penalty0 503--515, 1954.

\bibitem[Bello et~al.(2016)Bello, Pham, Le, Norouzi, and Bengio]{bello2016neural}
Bello, I., Pham, H., Le, Q.~V., Norouzi, M., and Bengio, S.
\newblock Neural combinatorial optimization with reinforcement learning.
\newblock \emph{arXiv preprint arXiv:1611.09940}, 2016.

\bibitem[Bi et~al.(2022)Bi, Ma, Wang, Cao, Chen, Sun, and Chee]{bi2022learning}
Bi, J., Ma, Y., Wang, J., Cao, Z., Chen, J., Sun, Y., and Chee, Y.~M.
\newblock Learning generalizable models for vehicle routing problems via knowledge distillation.
\newblock \emph{Advances in Neural Information Processing Systems}, 35:\penalty0 31226--31238, 2022.

\bibitem[Brohan et~al.(2022)Brohan, Brown, Carbajal, Chebotar, Dabis, Finn, Gopalakrishnan, Hausman, Herzog, Hsu, et~al.]{brohan2022rt}
Brohan, A., Brown, N., Carbajal, J., Chebotar, Y., Dabis, J., Finn, C., Gopalakrishnan, K., Hausman, K., Herzog, A., Hsu, J., et~al.
\newblock Rt-1: Robotics transformer for real-world control at scale.
\newblock \emph{arXiv preprint arXiv:2212.06817}, 2022.

\bibitem[Chen et~al.(2021)Chen, Lu, Rajeswaran, Lee, Grover, Laskin, Abbeel, Srinivas, and Mordatch]{chen2021decision}
Chen, L., Lu, K., Rajeswaran, A., Lee, K., Grover, A., Laskin, M., Abbeel, P., Srinivas, A., and Mordatch, I.
\newblock Decision transformer: Reinforcement learning via sequence modeling.
\newblock \emph{Advances in neural information processing systems}, 34:\penalty0 15084--15097, 2021.

\bibitem[Drakulic et~al.(2024{\natexlab{a}})Drakulic, Michel, and Andreoli]{drakulic2024goal}
Drakulic, D., Michel, S., and Andreoli, J.-M.
\newblock Goal: A generalist combinatorial optimization agent learning.
\newblock \emph{arXiv preprint arXiv:2406.15079}, 2024{\natexlab{a}}.

\bibitem[Drakulic et~al.(2024{\natexlab{b}})Drakulic, Michel, Mai, Sors, and Andreoli]{drakulic2024bq}
Drakulic, D., Michel, S., Mai, F., Sors, A., and Andreoli, J.-M.
\newblock Bq-nco: Bisimulation quotienting for efficient neural combinatorial optimization.
\newblock \emph{Advances in Neural Information Processing Systems}, 36, 2024{\natexlab{b}}.

\bibitem[Dubey et~al.(2024)Dubey, Jauhri, Pandey, Kadian, Al-Dahle, Letman, Mathur, Schelten, Yang, Fan, et~al.]{dubey2024llama}
Dubey, A., Jauhri, A., Pandey, A., Kadian, A., Al-Dahle, A., Letman, A., Mathur, A., Schelten, A., Yang, A., Fan, A., et~al.
\newblock The llama 3 herd of models.
\newblock \emph{arXiv preprint arXiv:2407.21783}, 2024.

\bibitem[Erd6s \& R{\'e}nyi(1960)Erd6s and R{\'e}nyi]{erd6s1960evolution}
Erd6s, P. and R{\'e}nyi, A.
\newblock On the evolution of random graphs.
\newblock \emph{Publ. Math. Inst. Hungar. Acad. Sci}, 5:\penalty0 17--61, 1960.

\bibitem[Fischetti et~al.(1998)Fischetti, Gonzalez, and Toth]{fischetti1998solving}
Fischetti, M., Gonzalez, J. J.~S., and Toth, P.
\newblock Solving the orienteering problem through branch-and-cut.
\newblock \emph{INFORMS Journal on Computing}, 10\penalty0 (2):\penalty0 133--148, 1998.

\bibitem[Golden et~al.(1987)Golden, Levy, and Vohra]{golden1987orienteering}
Golden, B.~L., Levy, L., and Vohra, R.
\newblock The orienteering problem.
\newblock \emph{Naval Research Logistics (NRL)}, 34\penalty0 (3):\penalty0 307--318, 1987.

\bibitem[Grinsztajn et~al.(2023)Grinsztajn, Furelos-Blanco, Surana, Bonnet, and Barrett]{grinsztajn2023winner}
Grinsztajn, N., Furelos-Blanco, D., Surana, S., Bonnet, C., and Barrett, T.
\newblock Winner takes it all: Training performant rl populations for combinatorial optimization.
\newblock \emph{Advances in Neural Information Processing Systems}, 36:\penalty0 48485--48509, 2023.

\bibitem[Gurobi~Optimization(2018)]{gurobi}
Gurobi~Optimization, L.
\newblock Gurobi optimizer reference manual, 2018.
\newblock URL \url{http://www.gurobi.com}.

\bibitem[Helsgaun(2017)]{helsgaun2017extension}
Helsgaun, K.
\newblock An extension of the lin-kernighan-helsgaun tsp solver for constrained traveling salesman and vehicle routing problems.
\newblock \emph{Roskilde: Roskilde University}, 12:\penalty0 966--980, 2017.

\bibitem[Janner et~al.(2021)Janner, Li, and Levine]{janner2021offline}
Janner, M., Li, Q., and Levine, S.
\newblock Offline reinforcement learning as one big sequence modeling problem.
\newblock \emph{Advances in neural information processing systems}, 34:\penalty0 1273--1286, 2021.

\bibitem[Kim et~al.(2022)Kim, Park, and Park]{kim2022sym}
Kim, M., Park, J., and Park, J.
\newblock Sym-nco: Leveraging symmetricity for neural combinatorial optimization.
\newblock \emph{Advances in Neural Information Processing Systems}, 35:\penalty0 1936--1949, 2022.

\bibitem[Kipf \& Welling(2016)Kipf and Welling]{kipf2016semi}
Kipf, T.~N. and Welling, M.
\newblock Semi-supervised classification with graph convolutional networks.
\newblock \emph{arXiv preprint arXiv:1609.02907}, 2016.

\bibitem[Kool et~al.(2018)Kool, Van~Hoof, and Welling]{kool2018attention}
Kool, W., Van~Hoof, H., and Welling, M.
\newblock Attention, learn to solve routing problems!
\newblock \emph{arXiv preprint arXiv:1803.08475}, 2018.

\bibitem[Kwon et~al.(2020)Kwon, Choo, Kim, Yoon, Gwon, and Min]{kwon2020pomo}
Kwon, Y.-D., Choo, J., Kim, B., Yoon, I., Gwon, Y., and Min, S.
\newblock Pomo: Policy optimization with multiple optima for reinforcement learning.
\newblock \emph{Advances in Neural Information Processing Systems}, 33:\penalty0 21188--21198, 2020.

\bibitem[Kwon et~al.(2021)Kwon, Choo, Yoon, Park, Park, and Gwon]{kwon2021matrix}
Kwon, Y.-D., Choo, J., Yoon, I., Park, M., Park, D., and Gwon, Y.
\newblock Matrix encoding networks for neural combinatorial optimization.
\newblock \emph{Advances in Neural Information Processing Systems}, 34:\penalty0 5138--5149, 2021.

\bibitem[Lamm et~al.(2017)Lamm, Sanders, Schulz, Strash, and Werneck]{kamis}
Lamm, S., Sanders, P., Schulz, C., Strash, D., and Werneck, R.~F.
\newblock Finding near-optimal independent sets at scale.
\newblock \emph{J. Heuristics}, 23\penalty0 (4):\penalty0 207--229, 2017.
\newblock \doi{10.1007/s10732-017-9337-x}.
\newblock URL \url{https://doi.org/10.1007/s10732-017-9337-x}.

\bibitem[Li et~al.(2021)Li, Yan, and Wu]{li2021learning}
Li, S., Yan, Z., and Wu, C.
\newblock Learning to delegate for large-scale vehicle routing.
\newblock \emph{Advances in Neural Information Processing Systems}, 34:\penalty0 26198--26211, 2021.

\bibitem[Liu et~al.(2024)Liu, Lin, Wang, Zhang, Xialiang, and Yuan]{liu2024multi}
Liu, F., Lin, X., Wang, Z., Zhang, Q., Xialiang, T., and Yuan, M.
\newblock Multi-task learning for routing problem with cross-problem zero-shot generalization.
\newblock In \emph{Proceedings of the 30th ACM SIGKDD Conference on Knowledge Discovery and Data Mining}, pp.\  1898--1908, 2024.

\bibitem[Nazari et~al.(2018)Nazari, Oroojlooy, Snyder, and Tak{\'a}c]{nazari2018reinforcement}
Nazari, M., Oroojlooy, A., Snyder, L., and Tak{\'a}c, M.
\newblock Reinforcement learning for solving the vehicle routing problem.
\newblock \emph{Advances in neural information processing systems}, 31, 2018.

\bibitem[Reed et~al.(2022)Reed, Zolna, Parisotto, Colmenarejo, Novikov, Barth-Maron, Gimenez, Sulsky, Kay, Springenberg, et~al.]{reed2022generalist}
Reed, S., Zolna, K., Parisotto, E., Colmenarejo, S.~G., Novikov, A., Barth-Maron, G., Gimenez, M., Sulsky, Y., Kay, J., Springenberg, J.~T., et~al.
\newblock A generalist agent.
\newblock \emph{arXiv preprint arXiv:2205.06175}, 2022.

\bibitem[Shafiullah et~al.(2022)Shafiullah, Cui, Altanzaya, and Pinto]{shafiullah2022behavior}
Shafiullah, N.~M., Cui, Z., Altanzaya, A.~A., and Pinto, L.
\newblock Behavior transformers: Cloning $ k $ modes with one stone.
\newblock \emph{Advances in neural information processing systems}, 35:\penalty0 22955--22968, 2022.

\bibitem[Singh \& Rizwanullah(2022)Singh and Rizwanullah]{singh2022combinatorial}
Singh, G. and Rizwanullah, M.
\newblock Combinatorial optimization of supply chain networks: A retrospective \& literature review.
\newblock \emph{Materials today: proceedings}, 62:\penalty0 1636--1642, 2022.

\bibitem[Sun \& Yang(2023)Sun and Yang]{sun2023difusco}
Sun, Z. and Yang, Y.
\newblock Difusco: Graph-based diffusion solvers for combinatorial optimization.
\newblock \emph{Advances in Neural Information Processing Systems}, 36:\penalty0 3706--3731, 2023.

\bibitem[Toth \& Vigo(2014)Toth and Vigo]{toth2014vehicle}
Toth, P. and Vigo, D.
\newblock \emph{Vehicle routing: problems, methods, and applications}.
\newblock SIAM, 2014.

\bibitem[Vaswani(2017)]{vaswani2017attention}
Vaswani, A.
\newblock Attention is all you need.
\newblock \emph{Advances in Neural Information Processing Systems}, 2017.

\bibitem[Vinyals et~al.(2015)Vinyals, Fortunato, and Jaitly]{vinyals2015pointer}
Vinyals, O., Fortunato, M., and Jaitly, N.
\newblock Pointer networks.
\newblock \emph{Advances in neural information processing systems}, 28, 2015.

\bibitem[Wen et~al.(2022)Wen, Wan, Zhou, Hou, Cao, Le, Chen, Tian, Zhang, and Wang]{wen2022realization}
Wen, Y., Wan, Z., Zhou, M., Hou, S., Cao, Z., Le, C., Chen, J., Tian, Z., Zhang, W., and Wang, J.
\newblock On realization of intelligent decision-making in the real world: A foundation decision model perspective.
\newblock \emph{arXiv preprint arXiv:2212.12669}, 2022.

\bibitem[Wolpert \& Macready(1997)Wolpert and Macready]{wolpert1997no}
Wolpert, D.~H. and Macready, W.~G.
\newblock No free lunch theorems for optimization.
\newblock \emph{IEEE transactions on evolutionary computation}, 1\penalty0 (1):\penalty0 67--82, 1997.

\bibitem[Zhang et~al.(2023)Zhang, Dai, Malkin, Courville, Bengio, and Pan]{zhang2023let}
Zhang, D., Dai, H., Malkin, N., Courville, A.~C., Bengio, Y., and Pan, L.
\newblock Let the flows tell: Solving graph combinatorial problems with gflownets.
\newblock \emph{Advances in neural information processing systems}, 36:\penalty0 11952--11969, 2023.

\bibitem[Zhao et~al.(2021)Zhao, Yu, and Xu]{zhao2021learning}
Zhao, H., Yu, Y., and Xu, K.
\newblock Learning efficient online 3d bin packing on packing configuration trees.
\newblock In \emph{International conference on learning representations}, 2021.

\bibitem[Zhou et~al.(2022)Zhou, Kumar, Finn, and Rajeswaran]{zhou2022policy}
Zhou, A., Kumar, V., Finn, C., and Rajeswaran, A.
\newblock Policy architectures for compositional generalization in control.
\newblock \emph{arXiv preprint arXiv:2203.05960}, 2022.

\bibitem[Zhou et~al.(2023)Zhou, Wu, Song, Cao, and Zhang]{zhou2023towards}
Zhou, J., Wu, Y., Song, W., Cao, Z., and Zhang, J.
\newblock Towards omni-generalizable neural methods for vehicle routing problems.
\newblock In \emph{International Conference on Machine Learning}, pp.\  42769--42789. PMLR, 2023.

\bibitem[Zong et~al.(2022)Zong, Wang, Wang, Zheng, and Li]{zong2022rbg}
Zong, Z., Wang, H., Wang, J., Zheng, M., and Li, Y.
\newblock Rbg: Hierarchically solving large-scale routing problems in logistic systems via reinforcement learning.
\newblock In \emph{Proceedings of the 28th ACM SIGKDD Conference on Knowledge Discovery and Data Mining}, pp.\  4648--4658, 2022.

\end{thebibliography}
\bibliographystyle{icml2025}


\appendix
\newpage

\section{Problem Details}
\label{problems}
In this section, we continue to introduce the implementation details on each CO problem. We use $N$ to denote either node, item or job amount, and $M$ to denote the total machine amount in FFSP. For each problem, we list the data generation scheme, the expert solver selection, the token (feature) design reference literature, prefix token designs and step token designs respectively. A brief summary is shown in Table~\ref{tab:problem summary}.

\begin{table*}[h]
    \caption{The summary of the evaluated CO problems, along with individual expert solver to collect trajectories, the prefix token length and the step state token length. $N$ denotes the number of nodes, items, or jobs, depending on the problem, and $M$ denotes the number of machines in the FFSP.}
    \label{tab:problem summary}
    \begin{center}
    \begin{tabular}{c|c|c|c}
    \hline
        \textbf{Problem} & \textbf{Expert Solver} &\textbf{Prefix-Token} & \textbf{State-Token}  \\ 
        \hline
        TSP  & LKH3~\citep{helsgaun2017extension}&$2N$&$2$ \\
        VRP  & LKH3~\citep{helsgaun2017extension}&$3N+2$&$3$  \\
        OP   & Gurobi~\citep{gurobi}& $3N+2$&$4$ \\
        PCTSP & ILS \footnote{https://github.com/jordanamecler/PCTSP} &$4N+2$&$3$\\
        SPCTSP & ILS &$4N+2$&$3$ \\
        Knapsack &Dynamic Programming& $2N$ &$1$ \\
        ATSP &LKH3~\citep{helsgaun2017extension}&$N\times N$&$N$\\
        MIS &Kamis~\cite{kamis}&$N \times N$ &$N$\\
        FFSP &MatNet~\citep{kwon2021matrix}&$N\times M$&$M+1$ \\
        3DBP &PCT~\cite{zhao2021learning}&-&$6\times(N+1)$ \\
    \hline
        
    \end{tabular}
    \end{center}
\end{table*}

\subsection{Traveling Salesman Problem (TSP)}
In the TSP, the objective is to should find the shortest route that visits each city exactly once and returns to the starting city. The objective is to minimize the total distance of the tour.

\textbf{Data Generation:}\quad We implement the dataset generation scheme described by \cite{kool2018attention}, for all TSP instances, the positions of $N$ nodes are uniformly randomly sampled in unit square.

\revise{
\textbf{Expert Solver:} \quad LKH~\citep{helsgaun2017extension}.

\textbf{Token (Feature) Design Reference:} \quad AM~\citep{kool2018attention}, POMO~\citep{kwon2020pomo}.

\textbf{Prefix Tokens:}\quad Coordinates of each city ($2N$ continuous values).

\textbf{Step State Tokens:} \quad  Coordinates of the current city ($2$ continuous values).

\textbf{Step Action Tokens:} \quad  The index of the city to visit next.
}

\subsection{Vehicle Routing Problem (VRP)}
In the Capacitated VRP~\citep{toth2014vehicle}, each city has a certain demand. The objective is to construct multiple routes with minimal a distance that all start and end at a given depot, where the total demands of cities within one route should not exceed the capacity limit. Except for the depot, each city should be visited exactly once.

\textbf{Data Generation.}\quad We implement the dataset described by \cite{nazari2018reinforcement}. Specifically, each city $i\in\{1,2,..,N\}$ has a demand $0<\delta_i\leq D$, where $D>0$ is the capacity of  the vehicle (route).  For each route $R_j$, the total demand of the cities along cannot exceed the vehicle's capacity, i.e. $\sum_{i\in R_j}\delta_i\leq D$.  For our experiments, We random sample the location coordinates of the depot and the cities within the unit square uniformly. The discrete demands are sampled uniformly from $\{1,2,...,9\}$ and the capacity is set to $D^{20}=30, D^{50}=40$. 

\revise{
\textbf{Expert Solver}: \quad LKH~\citep{helsgaun2017extension}.

\textbf{Token (Feature) Design Reference:} \quad AM~\citep{kool2018attention}, POMO~\citep{kwon2020pomo}.

\textbf{Prefix Tokens:}\quad Coordinates of depot and each city ($2(N+1)$ continuous values), demands of each city ($N$ continuous values).

\textbf{Step State Tokens:} \quad Coordinates of the current location ($2$ continuous values), current volume budget($1$ continuous value).

\textbf{Step Action Tokens:} \quad  The index of the location to visit next.
}

\subsection{Orienteering Problem (OP)} 

In the OP~\citep{golden1987orienteering}, each node is assigned with a specific prize. The objective is to construct a single tour that maximize the sum of prizes, starting and ending at a give depot. The tour does not have to include every node anymore, but need to be shorter than a length limit.

\textbf{Data Generation.}\quad We implement the data generation scheme by~\cite{fischetti1998solving,kool2018attention}. Specifically, The location coordinates of depot as well as $N$ node are random sampled uniformly in the unit square. To make the problem more challenging, we made the prize $p_i$ for each node $i$ proportional to its distance from the depot by setting them as:
\[p_{i}=1+\left[99 \cdot \frac{d_{0 i}}{\max _{j=1}^{n} d_{0 j}}\right], \hat{p}_{i}=\frac{p_{i}}{100}\]
where $d_{0i}$ is the distance from node $i$ to the depot. As for the length limit of the route, we set the fixed max length as $T^{20}=2$ and $T^{50}=3$, which makes the optimal number of access nodes different from instance to instance.

\revise{
\textbf{Expert Solver:}\quad Gurobi~\citep{gurobi}.

\textbf{Token (Feature) Design Reference:} \quad AM~\citep{kool2018attention}.

\textbf{Prefix Tokens:}\quad Coordinates of the depot and each city ($2(N+1)$ continuous values), prize of each city ($N$ continuous values).

\textbf{Step State Tokens:} \quad Coordinates of the current location ($2$ continuous values), total prize collected so far ($1$ continuous value), current length budget ($1$ continuous value).

\textbf{Step Action Tokens:} \quad  The index of the location to visit next.

}

\subsection{Prize Collecting TSP (PCTSP)} 
In the PCTSP~\citep{balas1989prize}, the sum of total prize is no longer a optimization objective, but a constraint. The objective is to minimize the total route length plus the sum of penalties of unvisited nodes which are given ahead, as well as collecting at least a minimal total prize.

\textbf{Data Generation.}\quad We implement the data generation scheme by ~\cite{kool2018attention}. Specifically, as the OP problem mentioned previously, the location coordinates of the depot and all nodes are randomly sampled uniformly within the unit square. For each node $i$, the associated prize $p_i$ and penalty $\beta_i$ need to be balanced carefully. If the penalty is too small, the choice of node is almost entirely determined by the total reward constraint; If the penalty is too large, all nodes are always accessed and the total reward constraint fails. Following the reference~\cite{kool2018attention}, we set the prize and penalty as:
\[t_{i} \sim \operatorname{Uniform}(0,1), \quad \rho_{i}=t_{i} \cdot \frac{4}{N} \]
\[\beta_{i} \sim \operatorname{Uniform}\left(0,3 \cdot \frac{K^{N}}{N}\right)\]
where $K^N$ is about half of the trajectory length of the TSP problem with $N$ cities, we roughly set it as $K^{20}=2, K^{50}=3$, and the minimum total prize is set to 1 for our experiments. 

\revise{
\textbf{Expert Solver:}\quad Iterated Local Search (ILS).

\textbf{Token (Feature) Design Reference:}\quad AM~\citep{kool2018attention}.

\textbf{Prefix Tokens:}\quad Coordinates of the depot and each city ($2(N+1)$ continuous values), prize of each city ($N$ continuous values), penalty of each city ($N$ continuous values).

\textbf{Step State Tokens:}\quad Coordinates of the current location ($2$ continuous values), prize-to-go to the minimum required total prize ($1$ continuous value).

\textbf{Step Action Tokens:} \quad  The index of the location to visit next.
}

\subsection{Stochastic PCTSP (SPCTSP)} In the SPCTSP, we show how UniCO performs when dealing with uncertainty. Compared to PCTSP,  the expected prize of each node is known before the optimization starts, while the real collected prize can only be revealed after visitation. 

\textbf{Data Generation.}\quad The data generation for SPCTSP is the sameas in PCTSP, except that we additionally generate the expected prize, which has the same distribution of the real prize.The expert solution algorithm is a modified version of ILS, where the tour is re-optimized iteratively, as suggested by~\cite{kool2018attention}.

\revise{
\textbf{Expert Solver:}\quad Modified Iterated Local Search (ILS) by suggested \cite{kool2018attention}.

\textbf{Token (Feature) Design Reference:}\quad AM~\citep{kool2018attention}.

\textbf{Prefix Tokens:}\quad Coordinates of the depot and each city ($2(N+1)$ continuous values), expected prize of each city ($N$ continuous values), penalty of each city ($N$ continuous values).

\textbf{Step State Tokens:}\quad Coordinates of the current location ($2$ continuous values), prize-to-go to the minimum required total prize ($1$ continuous value).

\textbf{Step Action Tokens:} \quad  The index of the location to visit next.
}

\subsection{Asymmetric TSP (ATSP)} In the ATSP, the distances between node pairs are no longer determined by Euclidean distances based on node coordinates. Considering a directed graph, the distances are no longer necessarily the same in both directions, and are given in an asymmetric cost matrix upfront. We show how our model performs when dealing with features of $O(N^2)$ complexity.

\textbf{Data Generation.}\quad We follow the same data generation scheme as we did for TSP instances. The cities are selected uniformly in a unit square but only adjacency matrix is visible to represent problem instance.

\revise{
\textbf{Expert Solver:}\quad LKH3~\citep{helsgaun2017extension}.

\textbf{Token (Feature) Design Reference:}\quad Raw feature usage.

\textbf{Prefix State Tokens:}\quad Adjacency matrix ($N \times N$ continuous values), serialized by rows.

\textbf{Step Tokens:}\quad The row of the current city in adjacency matrix ($N$ continuous values).

\textbf{Step Action Tokens:} \quad  The index of the city to visit next.
}

\subsection{Knapsack} In the Knapsack problem, a group of items with specific values and volumes are given. The objective is to maximize the total value of items selected without exceeding the total capacity. We designed the problem generation scheme manually and implemented the dynamic programming algorithm for trajectory collection.

\textbf{Data Generation.}\quad We implement a manually designed data generation scheme. Specifically, The values $v_i$ of each item $i\in\{1,2,...,N\}$ are randomly sampled as:
\[v_i \sim \operatorname{Uniform}(2,20)\]
To make the problem more challenging, items of higher value should have a larger volume. We further introduce some randomness and set the volume $k_i$ of item $i$ as:
\[k_i =(1+t)v_i\]
where $t\sim \operatorname{Uniform}(\{-0.5, 0.5\})$, which means we increase or decrease the volume of item $i$  uniformly and randomly.  we set the fixed total capacity as $T^{20}=30$ and $T^{50}=75$.

\revise{
\textbf{Expert Solver:}\quad Gurobi~\citep{gurobi}.

\textbf{Token (Feature) Design Reference:}\quad POMO~\citep{kwon2020pomo}.

\textbf{Prefix Tokens:}\quad Values of all items ($N$ discrete values), volumes of all items ($N$ discrete values).

\textbf{Step State Tokens:}\quad Current volume budget (1 discrete values).

\textbf{Step Action Tokens:} \quad  The index of the newly selected item.
}

\subsection{Maximum Independent Set (MIS)} In the MIS, an independent set is a set of vertices such that no two vertices in the set are adjacent. One should find the largest possible independent set in the graph, meaning it contains the most vertices among all possible independent sets. 

\textbf{Data Generation.}\quad
We follow the random graph generation scheme proposed by~\cite{erd6s1960evolution} , and directly implement the script provided by~\cite{sun2023difusco} to generate the graphs.

\revise{
\textbf{Expert Solver:}\quad Kamis~\citep{kamis}.

\textbf{Token (Feature) Design Reference:}\quad Raw feature usage.

\textbf{Prefix Tokens:}\quad Adjacency matrix ($N \times N$ discrete values), serialized by rows.

\textbf{Step State Tokens:}\quad Whether each node is selected, excluded or not decided yet. ($N$ discrete values).

\textbf{Step Action Tokens:} \quad  The index of the newly selected node.
}

\subsection{Flexible Flow Shop Problem (FFSP)}  
In the FFSP, $N$ jobs have to be processed in several stages with the same order. Each job in each stage can be handled by a machine from $M$ total machines. The time required for each job at different stages on different machines varies. Each machine can only process at most one job at the same time. The goal is to schedule all jobs so that they can be finished with a minimum of time. 

\textbf{Data Generation.}\quad
We directly adopt the data generation scheme and script provided by~\cite{kwon2021matrix}, where $N=20, M=12$. We further implement the corresponding MatNet as the only NCO expert solver in our experiments for trajectory generation. 

\revise{
\textbf{Expert Solver:}\quad MatNet~\citep{kwon2021matrix}.

\textbf{Token (Feature) Design Reference:}\quad MatNet~\citep{kwon2021matrix}.

\textbf{Prefix Tokens:}\quad Job durations in each stage on the corresponding machine of each job ($N \times M$ discrete values). 

\textbf{Step State Tokens:}\quad Job durations of the current machine ($M$ discrete values).

\textbf{Step Action Tokens:} \quad  The index of the newly selected job to the current machine, or halt.
}

\begin{figure*}[h]
    \centering
    \includegraphics[width=0.99\textwidth]{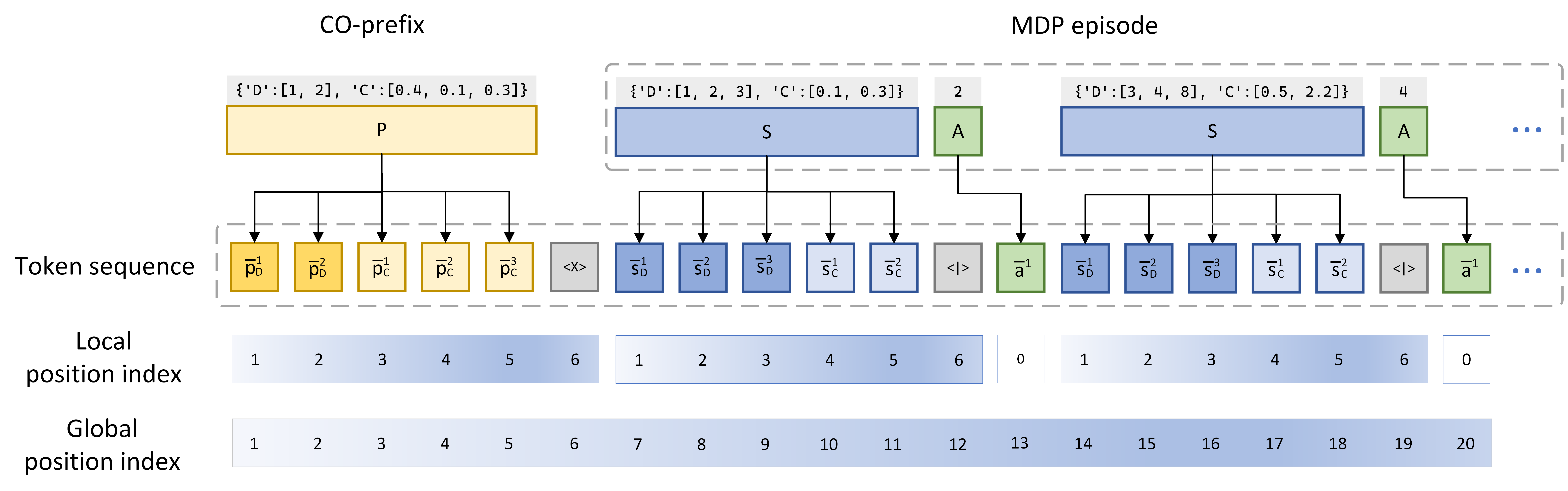}
    \vspace{-3mm}
    \caption{
        \revise{Tokenization illustration of CO-prefix and MDP sequence. 'D' includes all discrete values, and 'C' includes all continuous ones. }
    }
    \label{fig:tokenization}
    \vspace{-4mm}
\end{figure*}

\begin{figure*}[h]
\centering
\subfigure[$H>T$.]{
\includegraphics[width=0.45\textwidth]{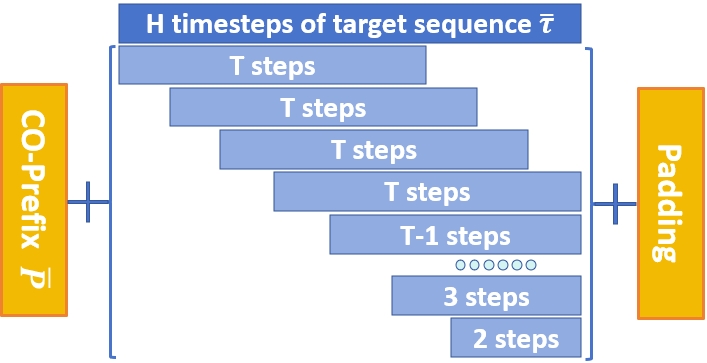}
}
\hfill
\subfigure[ $H<T$.]{
\includegraphics[width=0.45\textwidth]{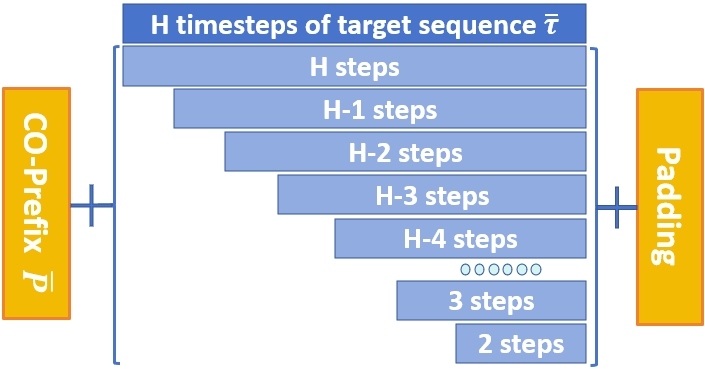}
}
\caption{
\revise{Trajectory collection illustration. Instead of directly using all transitions with $T$ steps of the original MDP episode, we collect subsequences and concatenate them to the prefix $\overline{P}$ as the trajectory data we use for training.}
}
\label{fig:traj-collection}
\vspace{-4mm}
\end{figure*}

\subsection{Online 3D Bin-packing (3DBP)}
In the 3DBP, a set of cuboid-shaped items should be packed. The upcoming items cannot be observed in advance, while only the current item to be packed is observable. The objective is to maximize the total utility rate of all boxes.

\textbf{Data Generation.}\quad
We directly adopt the data generation scheme and script provided by~\cite{zhao2021learning}. They use a novel hierarchical representation, packing configuration tree (PCT), to describe the state and action space during packing. We further implement the corresponding PCT approach as the only NCO expert solver in our experiments for trajectory generation.

\textbf{Expert Solver:}\quad PCT~\cite{zhao2021learning}.

\textbf{Token (Feature) Design Reference:}\quad PCT~\cite{zhao2021learning}.

\textbf{Prefix Tokens:} None, since 3DBP is fully dynamic and no preliminary information can be presented in advance. 

\textbf{Step State Tokens:}\quad PCT node configurations.

\textbf{Step Action Tokens:} \quad  The index of tree node to pack the next item in the current space.

\begin{figure*}[h]
    \centering
    \includegraphics[width=0.8\textwidth]{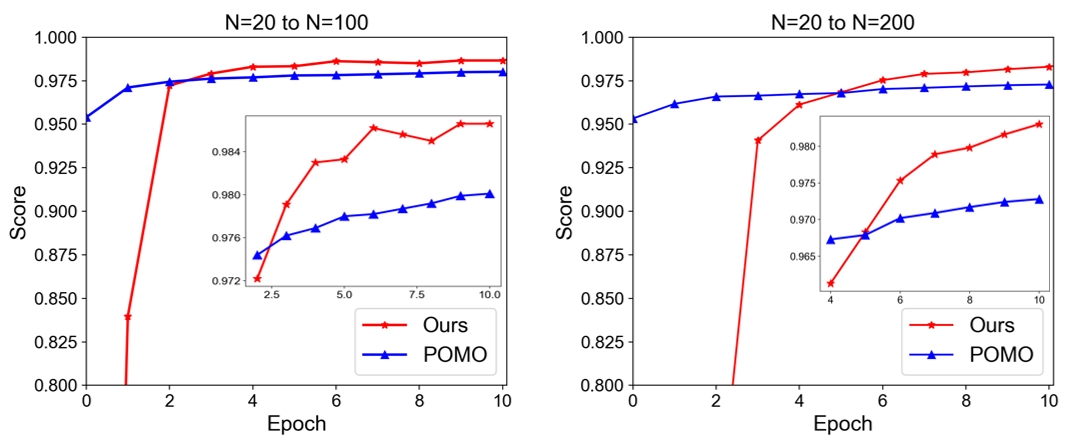}
    \caption{Results of finetuning UniCO trained with $N=20$ problems in Table~\ref{tab:main-result} to large scale TSP problem with $N=100$ and $N=200$. }
    \label{fig:scaleup}
\end{figure*}

\section{Tokenization and Trajectory Collection Details}
\label{appendix:tokenization}
In this section, we detail the tokenization and trajectory collection methods used in our model.

\subsection{Tokenization}
A complete trajectory sequence fed into our model consists of two components: the CO-prefix and the subsequent transition steps in the corresponding MDP episode, as illustrated in Figure~\ref{fig:tokenization}. Both raw CO-prefix $P$ and state $s_t$ at each step contain values that can be categorized into discrete and continuous types, as discussed in the previous section.   In most CO problems, the action representation is a discrete value. Both continuous and discrete values are flattened into a one-dimensional sequence and tokenized separately.

\begin{itemize}[leftmargin=*]
    \item As for continuous values, our goal is to discretize them and map them to unique token IDs. To achieve this, we use mu-law transformation to convert all values into a fixed range. The mu-law transformation is a common technique to handle continuous signals, ensuring that the values are transformed into a finite range suitable for tokenization. The formula for the mu-law transformation is:
    \begin{equation}
        F(x) = sgn(x)\frac{log(|x|\mu+1.0)}{log(M\mu+1.0)}
    \end{equation}
    where $M = 4$ and $\mu = 15$ in our experiments, and could be adjusted according to different data distribution.  The transformed values are further discretized via $N_{bin} = 1800$ bins, and mapped with token IDs of $\mathbb{Z}\in [200, 2000)$. 
    \item As for discrete values, we directly assign them with token IDs from the integer range $\mathbb{Z}\in [0, 200)$. All discrete values encountered in our previous experiments are strictly less than 200, ensuring that this range is sufficient to cover all discrete values in the data.
\end{itemize}

In addition to the discrete and continuous values, we also introduce two special tokens for separating key parts of the trajectory sequence. 
\begin{itemize}[leftmargin=*]
    \item Action Splitter: The token \texttt{<|>}, which separates the state tokens from the action tokens at each step, is assigned the token ID 2000.
    \item Prefix Splitter: The token \texttt{<X>}, which separates the CO-prefix from the subsequent MDP episode, is assigned the token ID 2001.
\end{itemize}

Once the tokens have been assigned, they are embedded into a continuous vector space using a lookup table. This embedding approach, where each token is mapped to a fixed-length vector, is consistent with the methods used in previous works such as ~\cite{reed2022generalist} and ~\cite{janner2021offline}. For position encoding, we employ a combination of both local and global position encodings. The local position encoding uses the local index within each step $\overline{\tau}_t$ or the prefix $\overline{P}$, while the global position encoding follows the traditional approach.

\subsection{Trajectory Collection and Data Augmentation}
In contrast to previous specialist NCO models, which typically use each raw problem instance only once during training or augment it based on symmetries of the CO problem~\citep{kool2018attention, kwon2020pomo}, UniCO employs a different data collection strategy. Each raw problem instance, along with its expert solution trajectory, can be used to generate multiple trajectory data for training, either complete or partial, as illustrated in Figure~\ref{fig:traj-collection}.

We set the target total token length $L$ ($L=1000$ in our main results) in advance, and compute the length of CO-prefix token length for each instance. The remaining token length, which will be allocated to the trajectory data $\overline{\tau}$, is determined by subtracting the CO-prefix token length from the target total token length $L$. The remaining token length corresponds to the maximum number of time steps $H$ in the target sequence $\overline{\tau}$.

Next, we use the total time steps $T$ from the complete MDP episode and clip subsequences from the original trajectory. If $H>T$, we clip subsequences with steps in the range of $[2, T]$. If $H<=T$, we clip subsequences with steps in the range of $[2, H]$. These subsequences are concatenated to the CO-prefix $\overline{P}$ to form a complete tokenized trajectory. It will be further padded to the target token length $L$, ensuring that each trajectory can be processed in parallel within a batch. The padded tokens are masked during computation so they do not affect model training.

This approach allows for significant data augmentation, as a single problem instance can generate multiple unique trajectories. Importantly, we do not restrict each trajectory to start from its very first time step during training. Instead, the model learns from the internal transitions between various steps in the trajectory, enhancing its ability to generalize across different stages of the solution process.

\begin{table*}
\centering

\caption{
\revise{
Performances comparison on different problem combinations. Three combinations are considered: all 9 problems, 6 routing problems and 3 non-routing problems. The best results are highlighted in bold.}
}
\label{tab:combination}
\begin{tabular}{c|cc|cc|cc}
\hline
         & \multicolumn{2}{c|}{All Problems}                     & \multicolumn{2}{c|}{Routing Problems}                 & \multicolumn{2}{c}{Non-Routing Problems}             \\
         & \multicolumn{1}{c}{Obj.} & \multicolumn{1}{c|}{Score} & \multicolumn{1}{c}{Obj.} & \multicolumn{1}{c|}{Score} & \multicolumn{1}{c}{Obj.} & \multicolumn{1}{c}{Score} \\
\hline\hline
TSP      & \textbf{3.87}            & \textbf{99.55\%}          & 4.03                     & 96.75\%                   & -                        & -                         \\
CVRP     & \textbf{6.66}            & \textbf{92.30\%}          & 6.89                     & 88.72\%                   & -                        & -                         \\
PCTSP    & \textbf{3.20}            & \textbf{99.34\%}          & 3.38                     & 96.25\%                   & -                        & -                         \\
OP       & \textbf{5.06}            & \textbf{90.72\%}          & 4.74                     & 80.02\%                   & -                        & -                         \\
SPCTSP   & \textbf{3.26}            & \textbf{100.84\%}         & 3.39                     & 98.55\%                   & -                        & -                         \\
ATSP     & 4.22                     & 94.43\%                   & \textbf{4.11}            & \textbf{95.80\%}          & -                        & -                         \\

Knapsack & \textbf{61.99}           & \textbf{92.62\%}          & -                        & -                         & 61.95                    & 92.47\%                   \\
MIS      & \textbf{10.35}           & \textbf{93.23\%}          & -                        & -                         & 10.28                    & 87.97\%                   \\
FFSP     & \textbf{29.10}           & \textbf{89.88\%}          & -                        & -                         & 29.11                    & 89.85\%     \\
\hline
\end{tabular}
\end{table*}

\begin{table*}
\caption{
\revise{Ablation study on different embedding dimensions. The best results are in bold. }
}
\label{tab:hyperparam}
\centering
\resizebox{\textwidth}{!}{
\begin{tabular}{c|cccccccccc}
\hline
                      & \multicolumn{2}{c}{h=128}         & \multicolumn{2}{c}{h=256}       & \multicolumn{2}{c}{h=512}        & \multicolumn{2}{c}{h=768}          & \multicolumn{2}{c}{h=1024}        \\
\multicolumn{1}{l|}{} & \multicolumn{2}{c}{\#params=2.7M} & \multicolumn{2}{c}{\#params=9M} & \multicolumn{2}{c}{\#params=34M} & \multicolumn{2}{c}{\#params=75M}   & \multicolumn{2}{c}{\#params=131M} \\
                      & Obj.            & Score           & Obj.           & Score          & Obj.            & Score          & Obj.            & Score            & Obj.           & Score            \\
\hline\hline
TSP                   & 6.82            & 94.44\%         & 6.02           & 98.37\%        & 5.94            & 98.78\%        & 5.96            & 98.66\%          & \textbf{5.92}  & \textbf{98.83\%} \\
CVRP                  & 12.55           & 89.13\%         & 11.75          & 93.12\%        & 11.59           & 93.87\%        & \textbf{11.59}  & \textbf{93.89\%} & 11.68          & 93.41\%          \\
OP                    & 11.22           & 60.07\%         & 15.18          & 89.41\%        & 15.55           & 92.16\%        & 15.37           & 90.76\%          & \textbf{15.61} & \textbf{92.62\%} \\
PCTSP                 & 5.56            & 93.30\%         & 4.91           & 97.33\%        & 4.77            & 98.20\%        & 4.81            & 98.00\%          & \textbf{4.71}  & \textbf{98.59\%} \\
Knapsack              & 140.14          & 69.94\%         & 160.28         & 97.69\%        & 160.28          & 97.65\%        & \textbf{160.49} & \textbf{97.96\%} & 160.36         & 97.80\%    \\     
\hline
\end{tabular}
}
\end{table*}

\section{Additional Results}

\subsection{Generalization to Larger Scales}
\label{scale-20-to-50}
In addition to evaluating UniCO on test sets of the same scale as the training set, we further analyze how well the model generalizes to larger-scale problems. To do so, we utilize the pre-trained model that was trained and reported in Table~\ref{tab:main-result} from Section 4. We then fine-tune this model on newly collected trajectory data for TSP with larger problem sizes: $N=100$ and $N=200$. The fine-tuning is performed for 10 epochs for each scale, and the results are compared with the POMO baseline~\citep{kwon2020pomo}.

The performance results are shown in Figure~\ref{fig:scaleup}, where we observe how the model adapts to larger problem sizes. Results demonstrate that POMO, as a specialist model, can be directly generalized to large scale problem even without finetuning. UniCO still requires finetuning steps to re-obtain problem solving ability. However, the necessary finetuning is fast. After only $3$ and $5$ epochs each, UniCO outperforms POMO already.  These results highlight the model's ability to scale effectively and provide valuable insights into the impact of fine-tuning on performance as the problem size increases.

\subsection{Analysis on Problem Combinations}
\label{prob-combination}
To better understand how the combination of different CO problems influences the performance of UniCO, we train the model on three distinct problem groups: (1) nine problems, (2) six routing problems, and (3) three non-routing problems. The performance results are evaluated via greedy decoding, and are shown in Table~\ref{tab:combination}.

Interestingly, we find that aggregating problem instances from structurally diverse problems can further boost the overall performance of the model. Except for ATSP, training on all nine problems together results in the best scores across all other problems compared to the other problem group combinations.

This observation demonstrates the effectiveness of UniCO trained on a diverse set of problems, as it can continuously improve its performance even as the data and problem types become more varied. This phenomenon aligns with findings from GATO~\citep{reed2022generalist}, where the model showed advantages when trained across different tasks, and we further confirm its applicability to combinatorial optimization problems. Our results provide compelling evidence that UniCO can generalize well across a wide range of CO problems.

\begin{table*}[!htbp] 
\caption{
\revise{Performance results with problem scales of 50. The best results among all learning-based models are underlined, and the best results among all unified models are in bold.}
}
\centering
\label{tab:N=50}
\resizebox{0.95\textwidth}{!}{
\begin{tabular}{c|cccc|cccc}
\hline\hline
                  & \multicolumn{4}{c|}{TSP}                                                  & \multicolumn{4}{c}{Knapsack}                                            \\
Method            & Obj.$\downarrow$ & Gap$\downarrow$ & Score$\uparrow$  & Time$\downarrow$ & Obj.$\uparrow$  & Gap$\downarrow$ & Score$\uparrow$  & Time$\downarrow$ \\
\hline
Random            & 26.08            & -               & 0.00\%           & (20s)            & 85.31           & -               & 0.00\%           & (1m)             \\
Expert            & 5.69             & 0.00\%          & 100.00\%         & (2h)             & 161.99          & 0.00\%          & 100.00\%         & (26m)            \\
POMO-single traj  & 5.73             & 0.70\%          & 99.80\%          & (37s)            & 161.04          & 0.59\%          & 98.76\%          & (1m)             \\
POMO              & \underline{5.70}&\underline{0.10\%}&\underline{99.97\%}&(1m)             & 161.87          & 0.08\%          & 99.84\%          & (2m)             \\
\hline
GATO/DB1-greedy   & 6.25             & 9.86\%          & 97.22\%          & (4h)             & 160.13          & 0.84\%          & 97.57\%          & (2h)             \\
GATO/DB1-sampling & 5.96             & 4.64\%          & 98.68\%          & (62h)            & 160.63          & 0.81\%          & 98.20\%          & (34h)            \\
UniCO-DR-greedy    & 5.99             & 5.27\%          & 98.53\%          & (4m)            & 160.36          & 1.01\%          & 97.80\%          & (2m)             \\
UniCO-greedy       & 5.93             & 4.38\%          & 98.77\%          & (4m)            & 160.68          & 0.81\%          & 98.20\%          & (2m)             \\
UniCO-sampling     & \textbf{5.78}    & \textbf{1.45\%} & \textbf{99.59\%} & (1h)             &\underline{\textbf{161.93}}&\underline{\textbf{0.04\%}}&\underline{\textbf{99.92\%}}&(29m)    \\
\hline\hline
                  & \multicolumn{4}{c|}{CVRP}                                                 & \multicolumn{4}{c}{OP}                                                  \\
Method            & Obj.$\downarrow$ & Gap$\downarrow$ & Score$\uparrow$  & Time$\downarrow$ & Obj.$\uparrow$  & Gap$\downarrow$ & Score$\uparrow$  & Time$\downarrow$ \\
\hline
Random            & 30.67            & -               & 0.00\%           & (1m)             & 3.14            & -               & 0.00\%           & (1m)             \\
Expert            & 10.35            & 0.00\%          & 100.00\%         & (12h)            & 16.59           & 0.00\%          & 100.00\%         & (5h)             \\
AM-greedy         & 10.97            & 5.88\%          & 97.00\%          & (20s)            & 16.01           & 3.34\%          & 95.84\%          & (11s)            \\
AM-sampling       &\underline{10.76} &\underline{3.79\%}&\underline{98.06\%}&(35m)           &\underline{16.55}&\underline{1.61\%}&\underline{98.01\%}&(12m)           \\
\hline
GATO/DB1-greedy   & 11.72            & 12.89\%         & 93.37\%          & (6h)             & 14.66           & 11.57\%         & 85.68\%          & (3h)             \\
GATO/DB1-sampling & 11.19            & 7.87\%          & 95.96\%          & (94h)            & 15.91           & 4.08\%          & 94.94\%          & (49h)            \\
UniCO-DR-greedy    & 11.68            & 12.82\%         & 93.42\%          & (5m)            & 15.38           & 7.29\%          & 91.00\%          & (4m)            \\
UniCO-greedy       & 11.61            & 12.14\%         & 93.77\%          & (5m)            & 15.49           & 6.64\%          & 91.77\%          & (4m)            \\
UniCO-sampling     & \textbf{11.06}   & \textbf{6.80\%} & \textbf{96.50\%} & (1h)    & \textbf{16.23}  & \textbf{2.07\%} & \textbf{97.44\%}          & (1h)             \\
\hline\hline
                  & \multicolumn{4}{c|}{PCTSP}                                                &                 &                 &                  &                 \\

Method            & Obj.$\downarrow$ & Gap$\downarrow$ & Score$\uparrow$  & Time$\downarrow$ &                 &                 &                  &                  \\
\cline{1-5}
Random            & 21.37            & -               & 0.00\%           & (1m)             &                 &                 &                  &                  \\
Expert            & 4.48             & 0.00\%          & 100.00\%         & (5h)             &                 &                 &                  &                  \\
AM-greedy         & 4.58             & 2.30\%          & 99.37\%          & (13s)            &                 &                 &                  &                  \\
AM-sampling       & \underline{4.53} &\underline{1.15\%}&\underline{99.69\%}&(22m)           &                 &                 &                  &                  \\
\cline{1-5}
GATO/DB1-greedy   & 4.92             & 9.89\%          & 97.27\%          & (4h)             &                 &                 &                  &                  \\
GATO/DB1-sampling & 4.63             & 3.23\%          & 99.11\%          & (65h)            &                 &                 &                  &                  \\
UniCO-DR-greedy    & 4.79             & 6.92\%          & 98.16\%          & (4m)            &                 &                 &                  &                  \\
UniCO-greedy       & 4.76             & 6.30\%          & 98.27\%          & (4m)            &                 &                 &                  &                  \\
UniCO-sampling     & \textbf{4.54}    & \textbf{1.36\%} & \textbf{99.63\%} & (1h)             &                 &                 &                  &                 \\

\end{tabular}
}
\end{table*}

\subsection{Ablation on Parameter Scales}
\label{ablation}
To better understand the effect of parameter scale on overall performance, we train several versions of our model with different parameter scales on five problems with $N=50$. Specifically, we focus on adjusting the width of the transformer backbone, i.e., the embedding dimensions. The results of these experiments are summarized in Table~\ref{tab:hyperparam}.

We observe that the performance of our model continues to improve as the total parameter scale increases. However, the rate of improvement gradually slows down when the total parameter scale reaches 75M and 131M, corresponding to embedding dimensions of 768 and 1024, respectively. Among these configurations, the model with 131M parameters outperforms the model with 75M parameters on 3 out of 5 problems.

While increasing the parameter scale generally improves performance, we find that further scaling the parameters beyond a certain point yields diminishing returns. This suggests that the current limitations are not solely related to parameter scale but may also be influenced by the number of problem types and the amount of data used for training. Moving forward, we aim to further explore how increasing the diversity of problem types and expanding the data size can enhance the scalability of our model, unlocking its full potential.

\subsection{Supplementary Performances on Larger Scales}
\label{scale-50}
In addition to the main results where $N=20$ for all problems, we further evaluate the performance of UniCO on larger problem scales. Specifically, we examine a problem scale of $N=50$ for five selected problems, and summarize the results in Table~\ref{tab:N=50}.

The results demonstrate that UniCO maintains consistent performance even as the problem scale increases from $N=20$ to $N=50$. Notably, our model outperforms the GATO/DB1 baseline and achieves performance comparable to that of single-model baselines. Our model even outperforms the POMO baseline on the Knapsack problem. These results underscore the robustness and scalability of UniCO, confirming that it is capable of handling problem instances with larger scales while maintaining high-quality performance across diverse CO problems.


\begin{figure}[h]
    \centering
    \includegraphics[width=0.43\textwidth]{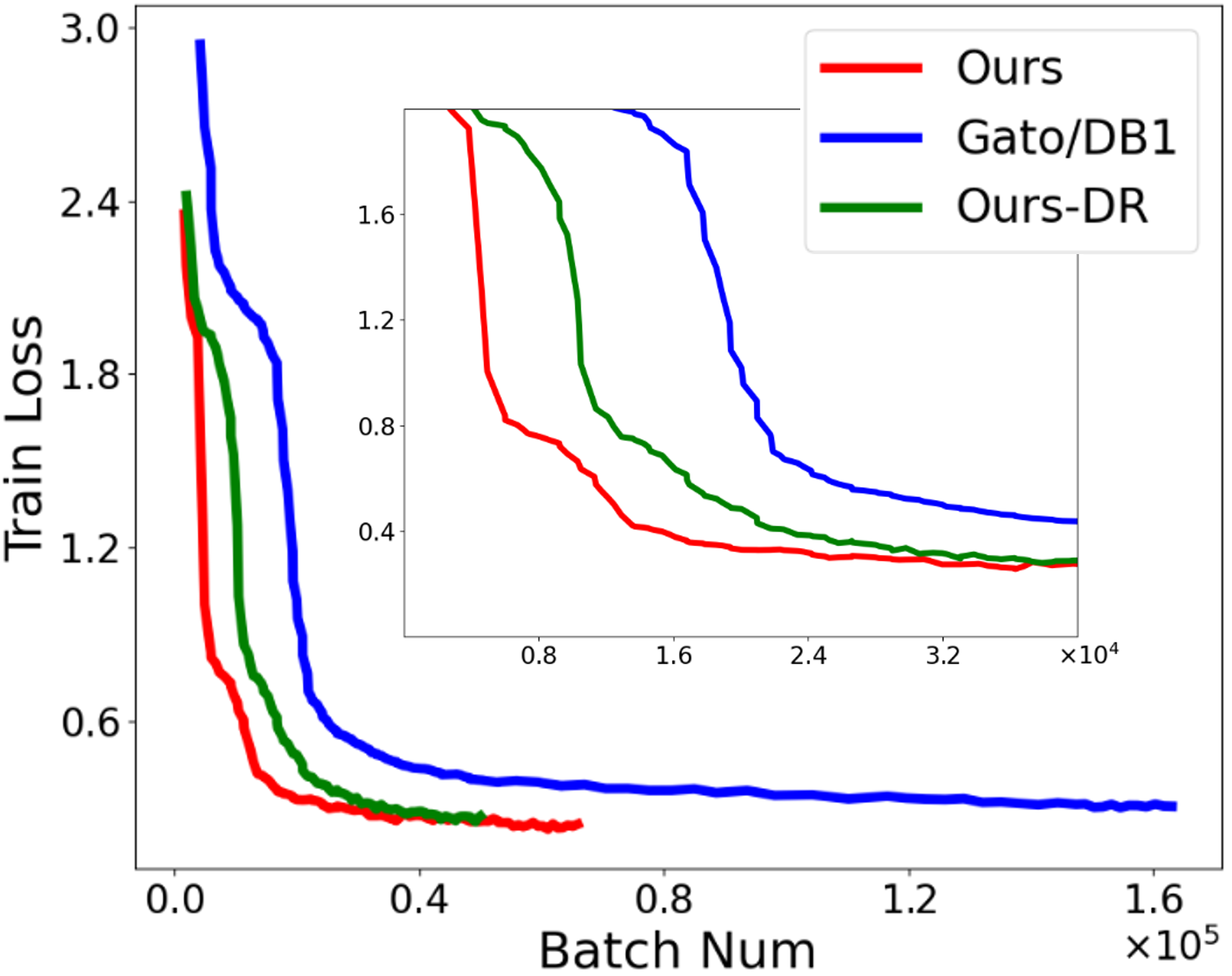}
    \caption{The losscurves along with total batch used of three models during training.}
    \label{fig:loss}
\end{figure}

\begin{table}[!h]
\caption{Implementation details.}
    \centering
    \resizebox{0.49\textwidth}{!}{
    \begin{tabular}{c|cc}
    \hline
                          Module&  Element  &  Detail  \\
     \hline\hline
       \multirow{5}{*}{System}  & OS    & Ubuntu 22.04.2\\
                                & CUDA  &   11.7\\
                                &Python &    3.11.4\\
                                &Pytorch&     2.0.1\\
                                &Device &      2*NVIDIA A100 80G\\
    \hline
        \multirow{22}{*}{Hyperparams}&Backbone    & Llama\\
                                 & Embedding dimension&768\\
                                 & Layer Num          &10\\
                                 & Q Head Num         &8\\
                                 & KV Head Num        &8\\
                                 & Total token length $L$&1000\\
                                 & RMS Norm epsilon   &1e-6\\
                                 & Weight Decay       &1e-4\\
                                 & Early Stopping Runs&6\\
                                 & M of $\mu$-law     &4\\
                                 &  $\mu$ of $\mu$-law&15\\
                                 & $[Min_d, Max_d)$     &$[0, 200)$\\
                                 & $[Min_d, Max_d)$     &$[200, 2000)$\\
                                 &Optimizer           &  AdamW\\
                                 &inital learning rate& 0\\
                                 & max learning rate  &2.5e-4\\
                                 &leanring rate warmup ratio& $5\%$\\
                                 & leanring rate decay ratio&$75\%$\\
                                 & leanring rate decay factor&10\\
                                 & leanring rate decay style&cosine\\
     \hline

    \end{tabular}
    }
    
    \label{tab:hyperparam}
\end{table}

\section{Evaluation Details and Training Process Reports.}
\label{appendix:implement}
We provide more implementation details for reproducibility. The hyperparameter used and environment settings are illustrated in Table~\ref{tab:hyperparam}. The detailed training process interms of loss is shown in Figure~\ref{fig:loss}. GATO/DB1 converges much slower than our model across all tasks.




\newpage
\begin{table*}[!htbp] 
\caption{
\revise{Performance results with problem scales of 50. The best results among all learning-based models are underlined, and the best results among all unified models are in bold.}
}
\centering
\label{tab:N=50}
\resizebox{0.95\textwidth}{!}{
\begin{tabular}{c|cccc|cccc}
\hline\hline
                  & \multicolumn{4}{c|}{TSP}                                                  & \multicolumn{4}{c}{Knapsack}                                            \\
Method            & Obj.$\downarrow$ & Gap$\downarrow$ & Score$\uparrow$  & Time$\downarrow$ & Obj.$\uparrow$  & Gap$\downarrow$ & Score$\uparrow$  & Time$\downarrow$ \\
\hline
Random            & 26.08            & -               & 0.00\%           & (20s)            & 85.31           & -               & 0.00\%           & (1m)             \\
Expert            & 5.69             & 0.00\%          & 100.00\%         & (2h)             & 161.99          & 0.00\%          & 100.00\%         & (26m)            \\
POMO-single traj  & 5.73             & 0.70\%          & 99.80\%          & (37s)            & 161.04          & 0.59\%          & 98.76\%          & (1m)             \\
POMO              & \underline{5.70}&\underline{0.10\%}&\underline{99.97\%}&(1m)             & 161.87          & 0.08\%          & 99.84\%          & (2m)             \\
\hline
GATO/DB1-greedy   & 6.25             & 9.86\%          & 97.22\%          & (4h)             & 160.13          & 0.84\%          & 97.57\%          & (2h)             \\
GATO/DB1-sampling & 5.96             & 4.64\%          & 98.68\%          & (62h)            & 160.63          & 0.81\%          & 98.20\%          & (34h)            \\
UniCO-DR-greedy    & 5.99             & 5.27\%          & 98.53\%          & (4m)            & 160.36          & 1.01\%          & 97.80\%          & (2m)             \\
UniCO-greedy       & 5.93             & 4.38\%          & 98.77\%          & (4m)            & 160.68          & 0.81\%          & 98.20\%          & (2m)             \\
UniCO-sampling     & \textbf{5.78}    & \textbf{1.45\%} & \textbf{99.59\%} & (1h)             &\underline{\textbf{161.93}}&\underline{\textbf{0.04\%}}&\underline{\textbf{99.92\%}}&(29m)    \\
\hline\hline
                  & \multicolumn{4}{c|}{CVRP}                                                 & \multicolumn{4}{c}{OP}                                                  \\
Method            & Obj.$\downarrow$ & Gap$\downarrow$ & Score$\uparrow$  & Time$\downarrow$ & Obj.$\uparrow$  & Gap$\downarrow$ & Score$\uparrow$  & Time$\downarrow$ \\
\hline
Random            & 30.67            & -               & 0.00\%           & (1m)             & 3.14            & -               & 0.00\%           & (1m)             \\
Expert            & 10.35            & 0.00\%          & 100.00\%         & (12h)            & 16.59           & 0.00\%          & 100.00\%         & (5h)             \\
AM-greedy         & 10.97            & 5.88\%          & 97.00\%          & (20s)            & 16.01           & 3.34\%          & 95.84\%          & (11s)            \\
AM-sampling       &\underline{10.76} &\underline{3.79\%}&\underline{98.06\%}&(35m)           &\underline{16.55}&\underline{1.61\%}&\underline{98.01\%}&(12m)           \\
\hline
GATO/DB1-greedy   & 11.72            & 12.89\%         & 93.37\%          & (6h)             & 14.66           & 11.57\%         & 85.68\%          & (3h)             \\
GATO/DB1-sampling & 11.19            & 7.87\%          & 95.96\%          & (94h)            & 15.91           & 4.08\%          & 94.94\%          & (49h)            \\
UniCO-DR-greedy    & 11.68            & 12.82\%         & 93.42\%          & (5m)            & 15.38           & 7.29\%          & 91.00\%          & (4m)            \\
UniCO-greedy       & 11.61            & 12.14\%         & 93.77\%          & (5m)            & 15.49           & 6.64\%          & 91.77\%          & (4m)            \\
UniCO-sampling     & \textbf{11.06}   & \textbf{6.80\%} & \textbf{96.50\%} & (1h)    & \textbf{16.23}  & \textbf{2.07\%} & \textbf{97.44\%}          & (1h)             \\
\hline\hline
                  & \multicolumn{4}{c|}{PCTSP}                                                &                 &                 &                  &                 \\

Method            & Obj.$\downarrow$ & Gap$\downarrow$ & Score$\uparrow$  & Time$\downarrow$ &                 &                 &                  &                  \\
\cline{1-5}
Random            & 21.37            & -               & 0.00\%           & (1m)             &                 &                 &                  &                  \\
Expert            & 4.48             & 0.00\%          & 100.00\%         & (5h)             &                 &                 &                  &                  \\
AM-greedy         & 4.58             & 2.30\%          & 99.37\%          & (13s)            &                 &                 &                  &                  \\
AM-sampling       & \underline{4.53} &\underline{1.15\%}&\underline{99.69\%}&(22m)           &                 &                 &                  &                  \\
\cline{1-5}
GATO/DB1-greedy   & 4.92             & 9.89\%          & 97.27\%          & (4h)             &                 &                 &                  &                  \\
GATO/DB1-sampling & 4.63             & 3.23\%          & 99.11\%          & (65h)            &                 &                 &                  &                  \\
UniCO-DR-greedy    & 4.79             & 6.92\%          & 98.16\%          & (4m)            &                 &                 &                  &                  \\
UniCO-greedy       & 4.76             & 6.30\%          & 98.27\%          & (4m)            &                 &                 &                  &                  \\
UniCO-sampling     & \textbf{4.54}    & \textbf{1.36\%} & \textbf{99.63\%} & (1h)             &                 &                 &                  &                 \\

\end{tabular}
}
\end{table*}

\end{document}